\documentclass[10pt,journal,compsoc]{IEEEtran}
\ifCLASSOPTIONcompsoc
  \usepackage[nocompress]{cite}
\else
  \usepackage{cite}
\fi

\usepackage{booktabs} 
\usepackage{epsfig}
\usepackage{epstopdf}
\usepackage{multirow}
\usepackage{algorithm}
\usepackage{algorithmic}
\usepackage{amssymb}
\usepackage{amsfonts}
\usepackage{amsmath}
\usepackage{graphicx}
\usepackage{subfigure}
\usepackage{cancel}
\usepackage{diagbox}
\usepackage{color}
\usepackage{bm}
\usepackage{pifont}
\usepackage{tikz}
\usepackage{makecell}
\usepackage{multirow}
\usepackage{hyperref}
\newcommand*{\circled}[1]{\lower.7ex\hbox{\tikz\draw (0pt, 0pt)%
    circle (.5em) node {\makebox[1em][c]{\small #1}};}}

\ifCLASSINFOpdf

\else

\fi

\begin{document}

\title{Revealing the Distributional Vulnerability of Discriminators by Implicit Generators}

\author{Zhilin Zhao,
        Longbing Cao
        and~Kun-Yu Lin
\IEEEcompsocitemizethanks{\IEEEcompsocthanksitem Zhilin Zhao and Longbing Cao are with the Data Science Lab, University of Technology Sydney, NSW 2007, Australia.\protect\\
E-mail: zhi-lin.zhao@student.uts.edu.au, longbing.cao@uts.edu.au
\IEEEcompsocthanksitem Kun-Yu Lin is with the School of Computer Science and Engineering, Sun Yat-sen University, Guangzhou, 515000, China.\protect\\
E-mail: linky5@mail2.sysu.edu.cn
}
}
\markboth{}
{Shell \MakeLowercase{\textit{et al.}}: Bare Demo of IEEEtran.cls for Computer Society Journals}

\IEEEtitleabstractindextext{%
\begin{abstract}
In deep neural learning, a discriminator trained on in-distribution (ID) samples may make high-confidence predictions on out-of-distribution (OOD) samples. This triggers a significant matter for robust, trustworthy and safe deep learning. The issue is primarily caused by the limited ID samples observable in training the discriminator when OOD samples are unavailable. We propose a general approach for \textit{fine-tuning discriminators by implicit generators} (FIG). FIG is grounded on information theory and applicable to standard discriminators without retraining. It improves the ability of a standard discriminator in distinguishing ID and OOD samples by generating and penalizing its specific OOD samples. According to the Shannon entropy, an energy-based implicit generator is inferred from a discriminator without extra training costs. Then, a Langevin dynamic sampler draws specific OOD samples for the implicit generator. Lastly, we design a regularizer fitting the design principle of the implicit generator to induce high entropy on those generated OOD samples. The experiments on different networks and datasets demonstrate that FIG achieves the state-of-the-art OOD detection performance.

\end{abstract}
}

\maketitle

\IEEEdisplaynontitleabstractindextext

%
\IEEEpeerreviewmaketitle

\IEEEraisesectionheading{\section{Introduction}
\label{sec:introduction}}
In deep learning, the discriminators built on deep neural networks (DNNs) demonstrate a significant generalization ability, conditioning on the independent and identically distributed (IID) assumption~\cite{RES:16,NN:19,CaoY022}. This assumes training and test samples must be drawn from the same distribution, i.e., \textit{in-distribution} (ID). However, this IID assumption may not hold as always \cite{Cao14}. A discriminator may make high-confidence predictions~\cite{HC:15} on those test samples~\cite{OOD:21} drawn from the distributions different from ID, i.e., \textit{out-of-distribution} (OOD) test samples~\cite{OOD:19}. This triggers the \textit{distributional vulnerability} of DNNs. The distributional vulnerability may result in various consequences for DNNs, including untrustworthy, unreliable, and unsafe results from DNNs. Specifically, a discriminator may fail to know whether an input is OOD. It may limit its adaption and cause serious technical and application issues, such as false identification, incorrect recognition, misclassification in vision tasks, language modeling, and false recommendation~\cite{BL:17,AT:17}. Therefore, addressing the distributional vulnerability of DNNs in distinguishing ID and OOD samples becomes a significant topic. This can ensure the robustness, trustworthiness, and safety of the learning results from DNNs~\cite{SF:16}.

One main cause of the high-confidence predictions of OOD samples in testing is the significant difference between ID and OOD samples~\cite{UN:17}. Specifically, in training with limited ID samples but no OOD samples, a discriminator learns to assign high-confidence predictions to the observed ID samples. This learning practice causes the \textit{distributional vulnerability} of neural networks, i.e., the trained discriminator may make high-confidence predictions for some OOD samples. This issue arises since the discriminator never learns to be sensitive to OOD samples during training. The issue may also be related to \textit{distributional uncertainty}~\cite{PN:18}, which, however, focuses on the low-confidence issue for ID samples. In contrast, distributional vulnerability causes some OOD samples to receive unexpected high-confidence predictions. Both distributional vulnerability and uncertainty arise due to the mismatch between training and test distributions.

One possible speculation is that the distributional vulnerability is discriminator-specific. This is because altering training data results in different network parameters~\cite{FIX:20}. Different network architectures generate different distributions of data representations~\cite{RE:13}. To reveal and patch the distributional vulnerability of a standard discriminator, one idea is to fine-tune the discriminator with OOD samples drawn from a specific OOD generator. This strategy makes the fine-tuned discriminator sensitive to distributional vulnerability. However, the relevant methods~\cite{MIXUP:18,DA:19} without considering data and network characteristics cannot generate specific OOD samples with semantic shift and high-confidence predictions. They define OOD samples according to prior knowledge without knowing distributional vulnerability, thereby misaddressing the distributional vulnerability of the given discriminator. Such methods cannot handle most high-confidence OOD samples specific to the discriminator.

In light of the above speculation, the following three research questions must be answered in exploring distributional vulnerability and improving the OOD sensitivity of a standard discriminator:
\begin{itemize}
    \item[(1)] \textit{How to design an OOD generator for a discriminator?} Training an extra generator for OOD samples is usually expensive~\cite{PIC:17,GLOW:18,GSN:14}. Also, the generator must be related to the given discriminator to produce specific OOD samples, making it harder to design the generator.
    \item[(2)] \textit{How to efficiently sample high-confidence OOD data from generators?} OOD samples with low confidence and misaddressing distributional vulnerability could mislead the fine-tuning process. Furthermore, the inefficiency of generating samples would result in a significant bottleneck in the fine-tuning.
    \item[(3)] \textit{How to apply the generated OOD samples to patch the vulnerability?} The discriminator should be regulated as OOD sensitive to prevent the corresponding generator from generating high-confidence OOD samples. This requires the contrastive regulation  to the design principle of the  generator.
\end{itemize}

We propose an approach for \textit{fine-tuning discriminators by implicit generators} (FIG) to address the above questions. For a standard discriminator learned from a training ID dataset, we create its \textit{implicit generator} without extra training\footnote{The same ID dataset is used in both learning the standard discriminator and fine-tuning it with the OOD samples generated by its corresponding implicit generator.}. The implicit generator has the same parameters as the discriminator and is used to generate OOD samples. The underlying insight is that FIG learns an OOD-sensitive discriminator by making it difficult to draw OOD samples from the corresponding implicit generator. Specifically, the implicit generator is proportional to the negative entropy of the output probabilities from the standard discriminator. The principle behind this construction method is that an OOD sample given by a high confidence prediction has a low entropy of class probabilities according to the Shannon entropy~\cite{MINE:18}. Since the constructed implicit generator is energy-based~\cite{EBM:06}, the samplers based on the Langevin dynamics~\cite{SGLD:11} can draw samples from energy-based models effectively to generate OOD samples. Then, according to the construction principle of the implicit generator, we penalize the OOD samples by flattening their class probabilities to make the discriminator sensitive to OOD samples.

Consequently, building on information theory, FIG is a general approach with theoretical ground and guarantees but without retraining. It creates an implicit generator corresponding to a given standard discriminator and refines the discriminator to improve its OOD sensitivity. Specifically, in FIG,
\begin{itemize}
  \item an implicit generator is proportional to the negative entropy of the output probabilities from a standard discriminator without extra training costs;
  \item a sampler based on the Langevin dynamics efficiently draws high-confidence OOD samples from the implicit generator; and
  \item a regularizer based on the design principle of the implicit generator encourages the high entropy of the generated OOD samples.
\end{itemize}

Our experiments demonstrate that FIG significantly improves the OOD sensitivity against the state-of-the-art methods with various network settings and on real image data. Section~\ref{sec:relatedwork} reviews the related work. Section~\ref{sec:algorithm} introduces FIG. The experiment results are shown in Section~\ref{sec:experiment}, and Section~\ref{sec:conclusion} offers some concluding remarks.

\section{Related Work}\label{sec:relatedwork}

Here, we briefly review the work related to this paper. We discuss both broadly related research and applications and focus on those topics specific to this paper on OOD detection.

\subsection{Broadly Related Research and Applications}
OOD detection~\cite{BL:17} aims to detect whether a test sample for a trained discriminator is drawn from an ID or OOD. The OOD detection performance evaluates the OOD sensitivity of discriminators. This OOD problem is broadly related to (1) various research topics, including non-IID learning \cite{CaoY022,Cao14,Cao-rs-eng}, anomaly and outlier detection~\cite{Pang21}, and open set recognition \cite{Scheirer12}; and (2) diverse learning tasks and applications such as gesture recognition, emotion recognition, semantic space, misclassification, language modeling, semantic segmentation, fault diagnosis, long-tailed recognition, and remote sensing \cite{MacedoRZOL22}.

First, the training ID and test OOD settings for OOD detection can be regarded as a special setting of non-IID learning \cite{Cao14,Cao-rs-eng,CaoY022}, which assumes data are not IID drawn from the same distribution. OOD detection can be regarded as focusing on learning the non-IIDness between training and test samplings. Second, anomaly and outlier detection \cite{Pang21,BahriSPS22} typically focuses on detecting those samples (rather than test ones) deviating from the majority in training, which could be on non-IID data \cite{XiangWRSDZ21}. OOD detection instead identifies those test samples deviating from the training ones in terms of distributional shift. Lastly, OOD detection differs from open set recognition~\cite{Scheirer12,Chen22} although neither focus on known classes. Open set recognition trains a discriminator to assign test samples to an extra unknown class that does not belong to any classes in the training set, while OOD detection does not involve extra unknown classes.

Specifically, the OOD detection problem in the deep learning context has attracted increasing and substantial attention in recent years \cite{Li-etal21,MacedoRZOL22}. The related research on OOD detection can be categorized into (1) designing effective OOD detectors, (2) designing generative models for OOD detection, and (3) enhancing the OOD sensitivity of discriminators. As these topics are core to DNNs and this paper, below, we extend our discussion on them.

\subsection{Out-of-distribution Detectors}
An OOD detector distinguishes ID and OOD test samples according to their predictions by a trained network (discriminator). The baseline method~\cite{BL:17} designs a threshold-based detector to distinguish ID and OOD samples, which involves their maximum probabilities represented by softmax outputs~\cite{KD:15}. It basically assumes that a trained discriminator tends to provide high-confidence predictions for ID samples. However, this assumption does not hold in general due to the distributional vulnerability, when OOD samples also have high softmax scores. To improve this baseline, an out-of-DIstribution detector for Neural networks (ODIN)~\cite{ODIN:18} adds negative adversarial perturbations to inputs to make ID and OOD samples distinguishable. Furthermore, ODIN applies temperature scaling to the softmax function and makes the trained discriminators more sensitive to OOD samples.

An OOD sample can be assigned with a high-confidence prediction because it is mapped to the feature representations of ID samples. This may cause feature collapse~\cite{FC:21}. Therefore, to improve the aforementioned softmax-based detectors, another set of detectors model the output distributions of various network layers. For example, MahaLanoBis (MLB)~\cite{MLB:18} combines the Mahalanobis distance calculation with input preprocessing to measure the OOD score according to the feature representations from different network layers. Based on ODIN and MLB, Deep Residual Flow (DRF)~\cite{DFR:20} leverages an expressive density model by normalizing flows to calculate the residual flows of each layer and each class for a test sample. Gram Matrix (GM)~\cite{GM:20} calculates the OOD score by identifying the feature correlations between activity patterns from all layers and the predicted class. However, detecting OOD samples without refining discriminators cannot resolve the vulnerability but makes the detection performance heavily dependent on the characteristics of the trained discriminators.

\subsection{Generative OOD Detection}
Another research line of detecting OOD samples in the test phase aims to learn an extra generative model according to the training ID samples. However, Nalisnick et al.~\cite{DGM:19} show that the likelihood alone of deep generative models, including flow-based models, VAEs, and PixelCNNs, fails to distinguish ID and OOD samples. Accordingly, Ren et al.~\cite{LR:19} show that the likelihood of auto-regressive models considering background statistics is sensitive to OOD samples. Serra et al.~\cite{LGM:20} provide an explanation of the failure of generative approaches, i.e., the input complexity has an excessive influence on the likelihoods. Therefore, they define the likelihood ratio as an OOD score based on the estimate of input complexity. The generative approaches are independent of discriminators and distinguish ID and OOD samples according to the likelihood-based scores. In contrast, our method improves the ability of discriminators to detect OOD samples from ID samples according to network outputs.


\subsection{Confidence Enhancement Methods}
To improve the OOD sensitivity of a discriminator by patching its distributional vulnerability, confidence enhancement methods~\cite{OOD:21} retrain or finetune the discriminator with extra knowledge about OOD samples. Some researchers apply real-world samples as OOD samples since they are drawn from the distributions different from the ID. For example, Outlier exposure~\cite{OE:19} randomly selects an OOD sample for each ID sample and enlarges the gap between the log probabilities of the pair of ID-OOD samples by a margin ranking loss. The prior network~\cite{PN:18} penalizes OOD samples by mapping their predicted distribution to a dense Dirichlet distribution in the Kullback-Leibler divergence. Bevandic et al.~\cite{SSS:19} propose a two-head model to predict a uniform distribution of OOD samples. Blum et al.~\cite{FH:19} separate ID and OOD samples by training a logistic regressor to aggregate the negative log-likelihoods of embeddings from all layers. Another line of research improves OOD sensitivity by modifying the training process and objective function without involving OOD samples in training. Built on ODIN, DeConf-C* (DCC*)~\cite{DCC:20} retrains a discriminator with an OOD scoring function according to the divisor structure of class probability confidence and searches for the adversarial perturbation magnitude with only ID samples.

Other methods aim to generate OOD samples and apply them to retrain a discriminator, which are mostly related to ours. They basically assume that OOD samples satisfy a normal or uniform distribution. However, this general assumption ignores data and network characteristics. Hence, the generated OOD samples cannot address the vulnerability of discriminators. Considering the data characteristics of OOD samples, MIXUP~\cite{MIXUP:18} trains a discriminator with samples obtained by linearly combining two randomly selected ID samples, where the weights are drawn from a beta distribution. When the weights are approximately equal to a half, the generated samples can be considered OOD because the target vector combining two one-hot vectors with two almost equal weights has low confidence. In addition, adversarial samples~\cite{AD:15} are applied to retain the discriminator, whose basic idea is to extend an input by pushing it to the decision boundary. Specifically, considering the network characteristics, adversarial samples~\cite{AD:15} are generated by back-propagating the gradient of the cross-entropy w.r.t. the input to a trained discriminator. Instead of manipulating data samples, joint confidence loss (JCL)~\cite{GO:18} extends the above idea to the distribution perspective. JCL adopts a model-specific GAN-based generator to produce samples on the low-density boundary of ID samples. It then encourages the target vectors of the generated samples to satisfy a uniform distribution. However, the JCL generated samples are at the border of the ID manifold, therefore the generated samples may not be OOD.

\section{FIG: Fine-tuning Discriminators by Implicit Generators} \label{sec:algorithm}
Our FIG method improves the OOD sensitivity of a standard discriminator by revealing and patching its distributional vulnerability. Specifically, for a standard discriminator learned from an ID dataset, FIG creates its implicit generator to generate specific OOD samples and applies a regularizer to fine-tune it with the generated OOD samples. The output class probabilities of the OOD sample assigned a high-confidence prediction have a low entropy. Accordingly, the implicit generator can be directly proportional to the negative entropy of the class probabilities from its corresponding standard discriminator. An implicit generator can then be induced from the discriminator. The specific OOD samples can be drawn from the implicit generator. According to the design principle of the implicit generator, FIG improves the OOD sensitivity of the standard discriminator by encouraging OOD samples to have large entropy.

We assume that ID samples $(\mathbf{x}_I, y_I)$ are IID drawn from an unknown distribution $p(\mathbf{x},y)$. $\mathbf{x} \in \mathbb{R}^D$ is a $D$-dimensional input, and $y \in \mathbb{R}$ is a label. $\mathcal{D}_I$ is the ID training dataset containing $N$ ID samples. As a typical machine learning setting, a $C$-class classification problem uses a parametric neural network $f_\theta: \mathbb{R}^D \rightarrow \mathbb{R}^C$ to map each input $\mathbf{x}$ to a $C$-dimensional output vector $(f_\theta(\mathbf{x},1),\ldots,f_\theta(\mathbf{x},C))$. A softmax output is applied to parameterize a categorical distribution for each output vector. Specifically, for class $y$, we estimate the probability $p(y|\mathbf{x})$ by:
\begin{equation}
\begin{aligned}
\label{eq:sm}
q_{\theta}(y | \mathbf{x}) & = \frac{\exp{f_{\theta}(\mathbf{x},y)}}{\sum_{y' \in [C]} \exp{f_{\theta}(\mathbf{x},y')}},
\end{aligned}
\end{equation}
and $q_{\theta}(y | \mathbf{x})$ is a standard discriminator with parameter $\theta$ learned from the ID training dataset $X_I$. In general, classification tasks learn parameter $\theta$ by maximizing the objective function $\mathbb{E}_{p(\mathbf{x},y)} \log q_{\theta}(y | \mathbf{x})$.
However, only limited ID samples following $p(\mathbf{x},y)$ are used to estimate the probability $q_{\theta}(y | \mathbf{x})$, which causes the vulnerability of the standard discriminator $q_{\theta}(y | \mathbf{x})$. Therefore, a critical step is to reveal where the vulnerability is before patching it.

\subsection{Implicit Generator}
A standard discriminator $q_{\theta}(y | \mathbf{x})$ learned from an ID dataset may provide high maximum softmax probabilities for some OOD samples due to distributional vulnerability. According to the definition of the Shannon entropy~\cite{MINE:18}, we know that the entropy values of high-confidence OOD samples are low. Accordingly, we define the entropy of a sample $\mathbf{x}$ as
\begin{equation}
\begin{aligned}
\label{eq:H}
H_{\theta,\mathbf{x}}(C) = -\sum_{y \in [C]} q_{\theta} (y | \mathbf{x}) \log q_{\theta} (y | \mathbf{x}).
\end{aligned}
\end{equation}
The range of $H_{\theta,\mathbf{x}}(C)$ is $(0, \log C]$. The joint energy-based model (JEM)~\cite{IG:18} contains a standard discriminator and a density model inferred by re-interpreting the logits of the discriminator. Inspired by JEM, we construct an implicit generator $q_{\theta}(\mathbf{x})$ for the discriminator $q_{\theta} (y | \mathbf{x})$ by assuming that the generator is proportional to the negative entropy, i.e.,
\begin{equation}
\begin{aligned}
\label{eq:entropy}
q_{\theta}(\mathbf{x}) & \propto - H_{\theta,\mathbf{x}}(C) + c \triangleq G(\mathbf{x})\\
\end{aligned}
\end{equation}
where a constant $c \geq \log C$ ($C \geq 0$) is added to ensure that the probability $q_{\theta}(\mathbf{x}) $ is proportional to a non-negative value. Based on the negative entropy, the samples drawn from $G$ should have high-confidence predictions but low entropy without necessarily having the same discriminator outputs as ID samples. Therefore, the low-entropy samples from $G$ are with distributional shift from training ID samples and have high-confidence predictions. This indicates that the low-entropy samples tend to be OOD. In JEM, the density model is inferred by re-interpreting the logits and marginalizing the label without constraints on the logit outputs. Therefore, the samples drawn from JEM are unnecessary for high-confidence predictions. JEM tends to generate samples similar to ID samples to ensure that they have the same logit outputs. In summary, the negative entropy enables an implicit generator to generate OOD samples, compared to JEM which generates ID samples.

However, sampling from $G$ is intractable because we cannot construct an analytic expression of the probability distribution $q_{\theta}(\mathbf{x})$ based on $G(\mathbf{x})$. Recall that the entropy value of a high-confidence OOD sample is expected to be low. Thus its $G(\mathbf{x})$ should be large. Accordingly, we specify a tractable probability distribution by exploring the upper bound of $G(\mathbf{x})$.

Assuming $h(\mathbf{x}) = \sum_{y' \in [C]} \exp{f_{\theta}(\mathbf{x},y')}$ and substituting Eq. (\ref{eq:sm}) and Eq. (\ref{eq:H}) into Eq. (\ref{eq:entropy}), we have
\begin{equation}
\begin{aligned}
\label{eq:q1}
G(\mathbf{x}) = &\frac{\sum_{y \in [C]} f_{\theta}(\mathbf{x},y) \exp{f_{\theta}(\mathbf{x},y)}}{h(\mathbf{x})} + \log \frac{\exp c}{h(\mathbf{x})}.\\
\end{aligned}
\end{equation}
To form a tractable bound, we set an upper bound on the second term of the last equality in Eq. (\ref{eq:q1}) using inequality: $\log(x) \leq \frac{x}{a} + \log(a) - 1$ for all $x,a \geq 0$. It is derived from the basic logarithm inequality $\log(1 + u) \leq u$, for $u > -1$ by assuming $u = \frac{x}{a} - 1$. We then obtain the following inequality,
\begin{equation}
\begin{aligned}
\label{eq:q2}
\log \frac{\exp c}{h(\mathbf{x})} & \leq \frac{\exp c}{h(\mathbf{x}) a(\mathbf{x})} + \log a(\mathbf{x}) - 1 = \frac{\exp (c - 1)}{h(\mathbf{x})}.\\
\end{aligned}
\end{equation}
We obtain the above equality by setting $a(\mathbf{x})$ as Euler's number $e$ because the inequality holds for any choice of $a(\mathbf{x}) \geq 0$. Substituting Eq. (\ref{eq:q2}) into Eq. (\ref{eq:q1}), we have
\begin{equation}
\begin{aligned}
\label{eq:G0}
& G(\mathbf{x}) \leq \frac{\sum_{y \in [C]} f_{\theta}(\mathbf{x},y) \exp{f_{\theta}(\mathbf{x},y)} + \exp (c - 1)}{h(\mathbf{x})} \\
= & \left[ \exp\left(  \underbrace{ \log \frac{\sum_{y \in [C]} \exp{f_{\theta}(\mathbf{x},y)}}{\sum_{y \in [C]} f_{\theta}(\mathbf{x},y) \exp{f_{\theta}(\mathbf{x},y)} + \exp (c - 1)}}_{\triangleq A(\mathbf{x})} \right) \right]^{-1}. \\
\end{aligned}
\end{equation}
To further obtain a tractable bound of $G(\mathbf{x})$, we need a lower bound on $A(\mathbf{x})$. According to the Jensen's inequality and inequality~\cite{VB:19} $\frac{x}{x + 1} \leq \log (1 + x) \leq x$ for all $x \geq -1$, respectively:
\begin{equation}
\begin{aligned}
\label{eq:A1}
\log \sum_{y \in [C]} \exp{f_{\theta}(\mathbf{x},y)} \geq \sum_{y \in [C]} f_{\theta}(\mathbf{x},y),
\end{aligned}
\end{equation}
and
\begin{equation}
\begin{aligned}
\label{eq:A2}
& \log \left( \sum_{y \in [C]} f_{\theta}(\mathbf{x},y) \exp{f_{\theta}(\mathbf{x},y)}  + \exp (c - 1) \right) \\
\leq & \sum_{y \in [C]} f_{\theta}(\mathbf{x},y) \exp{f_{\theta}(\mathbf{x},y)}  - \exp (c - 1) + 1.
\end{aligned}
\end{equation}
Substituting Eq. (\ref{eq:A1}) and Eq. (\ref{eq:A2}) into $A(\mathbf{x})$, we have
\begin{equation}
\begin{aligned}
\label{eq:A}
A(\mathbf{x}) \geq E_{\theta} (\mathbf{x}) -  \left( 1 - \exp (c - 1) \right).\\
\end{aligned}
\end{equation}
where
\begin{equation}
E_{\theta} (\mathbf{x}) \triangleq \sum_{y \in [C]} f_{\theta}(\mathbf{x},y) \left( 1 - \exp{f_{\theta}(\mathbf{x},y)} \right),
\end{equation}
is known as an energy function. It represents the state of $\mathbf{x}$ by mapping it to a scalar. Therefore, we obtain the upper bound of $G$ by substituting Eq. (\ref{eq:A}) into Eq. (\ref{eq:G0}):
\begin{equation}
\begin{aligned}
\label{eq:G}
G(\mathbf{x}) \leq & \exp \left( - E_{\theta}(\mathbf{x}) + \left( 1 - \exp (c - 1) \right) \right) \\
= & \exp\left(- E_{\theta}(\mathbf{x}) \right) \cdot \exp\left(\exp\left(c -1\right) - 1\right) \\
= & \frac{\exp\left( - E_{\theta}(\mathbf{x}) \right)}{\int \exp\left( - E_{\theta}(\mathbf{x}') \right) \mathrm{d} \mathbf{x}'} \cdot c',
\end{aligned}
\end{equation}
where $\int \exp\left( - E_{\theta}(\mathbf{x}') \right) \mathrm{d} \mathbf{x}'$ is a normalizing constant and
\begin{align}
c' = \int \exp\left( - E_{\theta}(\mathbf{x}') \right) \mathrm{d} \mathbf{x}' \cdot \exp\left(\exp\left(c -1\right) - 1\right)
\end{align}
is a constant, which is greater than or equal to zero and is independent of $\mathbf{x}$. Recall that $q_{\theta}(\mathbf{x}) \propto G(\mathbf{x})$, instead of directly solving $G(\mathbf{x})$ which is intractable, we take a tractable $q_{\theta}(\mathbf{x})$ by dropping the constant $c'$ according to the upper bound Eq. (\ref{eq:G}). This results in:
\begin{equation}
\begin{aligned}
\label{eq:Q}
q_{\theta}(\mathbf{x}) \propto \frac{\exp\left( - E_{\theta}(\mathbf{x}) \right)}{\int \exp\left( - E_{\theta}(\mathbf{x}') \right) \mathrm{d} \mathbf{x}'}.
\end{aligned}
\end{equation}
Therefore, we obtain the generator $q_{\theta}(\mathbf{x})$ from the given discriminator $q_{\theta}(y | \mathbf{x})$ without retraining. $q_{\theta}(\mathbf{x})$ has the same parameter $\theta$ as $q_{\theta}(y | \mathbf{x})$. Thus, $q_{\theta}(\mathbf{x})$ is the implicit generator of the standard discriminator $q_{\theta}(y | \mathbf{x})$.

\subsection{Langevin Dynamic Sampler}
We cannot easily draw samples from $q_{\theta}(\mathbf{x})$ because we do not have an analytical expression for $q_{\theta}(\mathbf{x})$. The analytical expression needs to integrate $\int \exp\left( - E_{\theta}(\mathbf{x}') \right) \mathrm{d} \mathbf{x}'$ with respect to $\mathbf{x}'$. However, $q_{\theta}(\mathbf{x})$ is an energy-based generative model~\cite{EBM:06}. $E_{\theta}(\mathbf{x})$ is the energy function. Relying on Markov chain Monte Carlo (MCMC)~\cite{MCMC:17} methods, random walk or Gibbs sampling~\cite{GI:02} can be applied, but both have long mixing time. The Langevin dynamics~\cite{SGLD:11} uses the gradient of the energy function. It can solve this challenge by drawing high-dimensional samples efficiently for energy-based models. Following the sampling method for energy-based models \cite{IG:19}, we apply the Langevin dynamic sampler (LDS) for the implicit generator $q_{\theta}(\mathbf{x})$ and have
\begin{equation}
\begin{aligned}
\label{eq:process}
 & \widetilde{\mathbf{x}}_t = \widetilde{\mathbf{x}}_{t - 1} - \frac{\epsilon_t}{2} \nabla_\mathbf{x} E_{\theta}(\widetilde{\mathbf{x}}_{t - 1}) + \mathbf{z}_t, \\
 & \mathbf{z}_t \sim \mathcal{N}(0, \epsilon_t \cdot \mathbf{I}), \\
 & \widetilde{\mathbf{x}}_0 \sim p_0(\mathbf{x}),
\end{aligned}
\end{equation}
$p_0(\mathbf{x})$ is an uniform distribution $\mathcal{U}(-1,1)$. $\epsilon$ is a decayed step-size. $\mathbf{I}$ is an identity matrix. When the number of iterations $T$ becomes infinite and the step-size $\epsilon_t$ is close to zero, the theoretical results provided by Welling and Teh~\cite{SGLD:11} guarantee that $\widetilde{\mathbf{x}}_T$ is a sample generated from the distribution defined by the energy function, that is
\begin{equation}
\begin{aligned}
\label{eq:tt}
\widetilde{\mathbf{x}}_T & \approx \widetilde{\mathbf{x}} \sim q_{\theta}(\mathbf{x})  (\epsilon_t \rightarrow 0 \, \text{and} \, T \rightarrow \infty).
\end{aligned}
\end{equation}
According to Eq.~(\ref{eq:process}), the optimization of Langevin dynamics can be treated as finding a local optimal solution $\widetilde{\mathbf{x}}_T$ from a posterior distribution that minimizes the energy function $E_{\theta}(\mathbf{x})$. In this aspect, the Langevin dynamics is similar to stochastic gradient descent~\cite{ML:14}. However, one clear difference between them is that Langevin dynamics injects noise into the parameter updates. The noise ensures that the trajectory of the parameters will converge to the whole posterior distribution rather than just the point with the highest posterior probability. Beyond that, Langevin dynamics is significantly different from the projected gradient descent method~\cite{PGD:18} applied in adversarial learning. The former finds a local optimal point while the latter finds a saddle point for a min-max problem.

Note that, $E_{\theta}(\mathbf{x})$ can be infinite because the large output value $f_{\theta}(\mathbf{x},y)$ in $E_{\theta}(\mathbf{x})$ can lead to the infinite exponential value $\exp f_{\theta}(\mathbf{x},y)$. Hence, instead of using $E_{\theta}(\mathbf{x})$ to construct the implicit generator $q_{\theta}(\mathbf{x})$, we apply the modified version
\begin{equation}
\begin{aligned}
\label{eq:mef}
\widehat{E}_{\theta}(\mathbf{x}) = \sum_{y \in [C]} \frac{f_{\theta}(\mathbf{x},y)}{c} \left( 1 - \exp{\frac{f_{\theta}(\mathbf{x},y)}{c}} \right)
\end{aligned}
\end{equation}
where $c$ is a constant to narrow $f_{\theta}(\mathbf{x},y)$. The step-size $\epsilon_t$ is updated by
\begin{equation}
\begin{aligned}
\epsilon_{t + L} = \epsilon_t \cdot \gamma,
\end{aligned}
\end{equation}
where $\gamma$ is the decay rate, and $L$ is the decay period. Following the process of generating adversarial samples~\cite{AD:15}, only the direction information is adopted to update the generated sample. This trick can improve sampling efficiency and avoid exploding gradients. We also clip the updated samples to the range $[-1,1]$ to ensure the consistency with the normalized input samples, i.e.,
\begin{equation}
\begin{aligned}
\label{eq:sample}
\widetilde{\mathbf{x}}_t = \text{clip}\left(\widetilde{\mathbf{x}}_{t - 1} - \frac{\epsilon_t}{2} \text{sign}(\nabla_\mathbf{x} \widehat{E}_{\theta}(\widetilde{\mathbf{x}}_{t - 1})) + \mathbf{z}_t, -1, 1 \right).
\end{aligned}
\end{equation}

In practice, it is impossible and unnecessary to generate low-entropy samples by following the theoretical results proposed by Welling and Teh~\cite{SGLD:11}, i.e., by running Eq. (\ref{eq:sample}) for an unlimited number of times, as shown in Eq. (\ref{eq:tt}). The prediction confidence for low-entropy samples is expected to be low. Only high-confidence samples should be penalized to patch the distributional vulnerability. We thus only need to explore the high-confidence samples to reveal the distributional vulnerability and ignore the low-confidence samples. Therefore, we repeat the iteration in Eq. (\ref{eq:sample}) until the confidence score of a generated low-entropy sample converges. LDS for generating a low-entropy sample is summarized in Algorithm~\ref{alg:sampler}.

Note that we find the confidence score of a generated sample can converge for a small maximum iteration $T \in [10,100]$. This means that the step size $\epsilon_t$ does not need to change to pursue low-entropy samples. However, more iterations are required to generate visually meaningful images for visualization, where the step-size should be adjusted to guarantee convergence. We further discuss the number of iterations $T$ on the visualization experiments.
\begin{algorithm}[t]
    \caption{Langevin Dynamic Sampler (LDS)}
    \label{alg:sampler}
    \begin{algorithmic}[1]
    \STATE {\textbf{Input:} discriminator $q_{\theta}(y|\mathbf{x})$}
    \STATE Initialize $\widetilde{\mathbf{x}}_0 \sim \mathcal{U}(-1,1)$, $\epsilon_0$, $\gamma$, $L$
    \WHILE{not converged}
        \STATE $\mathbf{z}_t \sim \mathcal{N}(0, \epsilon_t \cdot \mathbf{I})$
        \STATE $\widetilde{\mathbf{x}}_t = \text{clip} (\widetilde{\mathbf{x}}_{t - 1} - \frac{\epsilon_t}{2} \text{sign}(\nabla_\mathbf{x} \widehat{E}_{\theta}(\widetilde{\mathbf{x}}_{t - 1})) + \mathbf{z}_t, -1, 1  )$
        \STATE $\epsilon_{t + L} = \epsilon_t \cdot \gamma$
    \ENDWHILE
    \STATE {\bfseries Output:} $\widetilde{\mathbf{x}}_t$
\end{algorithmic}
\end{algorithm}

Accordingly, we can reveal the distributional vulnerability of a given discriminator by sampling discriminator-specific OOD samples in terms of Eq. (\ref{eq:sample}). We expect that all OOD samples have low prediction confidence, while the existence of vulnerability makes it impossible. Note that ID samples also have high confidence predictions and low entropy. Based on the assumption for implicit generators which are proportional to negative entropy, the generated samples per Eq. (\ref{eq:sample}) can also be ID. We assume that most drawn samples are OOD because ID samples have limited classes while the generated samples are diverse. The implicit generator inferred from a standard discriminator aims to reveal its distributional vulnerability. It differs from the generative adversarial network~\cite{GAN:14}, which learns from training ID samples to generate real-world objects. Therefore, the low-entropy samples drawn from the implicit generator are required to have high confidence and differ from ID samples. Such samples, however, unnecessarily correspond to real-world objects. According to the negative entropy principle of implicit generators, the generated samples have high-confidence predictions. They are not necessary to satisfy the same distribution as training ID samples. Therefore, the generated samples are almost OOD. Furthermore, even if some generated samples follow the ID, they will not affect the patching of the distributional vulnerability, as discussed in Section \ref{sec:cp}.

\subsection{Confidence Penalty on Out-of-distribution Samples}\label{sec:cp}
Due to distributional vulnerability, the prediction confidence on OOD samples by a standard discriminator could be unexpectedly high. Therefore, the low-entropy samples receiving high-confidence predictions by an implicit generator can reveal the distributional vulnerability of the corresponding standard discriminator. Accordingly, to patch this vulnerability, we penalize the low-entropy samples by flattening their class probabilities. Because the implicit generator depends on the corresponding standard discriminator, we improve the OOD sensitivity of the standard discriminator by making it difficult for the corresponding implicit generator to generate high-confidence OOD samples. Specifically, an implicit generator is proportional to negative entropy to ensure that the generated samples have high confidence predictions. We correspondingly penalize these low-entropy samples by encouraging them to have large entropy, i.e.,
\begin{equation}
\begin{aligned}
\label{eq:ob1}
& \max_{\theta} \mathbb{E}_{p(\mathbf{x},y)} \log q_{\theta}(y | \mathbf{x}) - \mathbb{E}_{q_{\theta}(\mathbf{x})} \sum_{y' \in [C]} q_{\theta} (y' | \mathbf{x}) \log q_{\theta} (y' | \mathbf{x}).\\
\end{aligned}
\end{equation}

After updating parameter $\theta$, we obtain an updated discriminator. Also, we can derive a new implicit generator and obtain the newly generated samples for the next iteration. We learn parameter $\theta$ iteratively until the implicit generator barely generates the samples with high confidence predictions. Note that the generated samples could be ID and OOD. Although some of the generated ID samples are also encouraged to have flat class probabilities, the rest of the generated OOD samples can still patch the vulnerability, and the dominated cross-entropy maintains the classification ability of the discriminator. In an extreme case where all generated samples are ID, the objective function Eq. (\ref{eq:ob1}) degenerates into the neural network confidence penalty method~\cite{CP:17}, which has empirically been demonstrated to improve the generalization ability.

We apply the stochastic gradient descent (SGD)~\cite{ML:14} optimization algorithm to estimate the gradient of the objective function Eq. (\ref{eq:ob1}). For the ID training dataset $\mathcal{D}_I$ containing $N$ ID samples, we draw $N \cdot K$ samples from the implicit generator $q_{\theta}(\mathbf{x})$ to construct the generated dataset $\mathcal{D}_O$. $K \in [0,1]$ is a hyper-parameter indicating the percentage of the generated low-entropy samples. In line with the idea of Monte Carlo~\cite{MC:19}, we estimate the objective function Eq. (\ref{eq:ob1}) by
\begin{equation}
\begin{aligned}
\label{eq:ob2}
\mathcal{L}(\theta) = & \frac{1}{N} \sum_{(\mathbf{x}_I, y_I) \in \mathcal{D}_I} \log q_{\theta}(y_I | \mathbf{x}_I) \\
& - \frac{1}{N \cdot K} \sum_{y' \in [C]} \sum_{\mathbf{x}_O \in \mathcal{D}_O} q_{\theta} (y' | \mathbf{x}_O) \log q_{\theta} (y' | \mathbf{x}_O).\\
\end{aligned}
\end{equation}

Algorithm~\ref{alg:FIG} summarizes the process of FIG to patch the distributional vulnerability of a standard discriminator with the low-entropy samples generated by its corresponding implicit generator.
\begin{algorithm}[t]
    \caption{FIG: Fine-tuning Discriminators by Implicit Generators}
    \label{alg:FIG}
    \begin{algorithmic}[1]
    \STATE {\textbf{Input:} standard discriminator $q_{\theta}(y|\mathbf{x})$, \\
    \quad \quad \quad low-entropy percentage $K$, \\
    \quad \quad \quad learning rate $\mu$}
    \REPEAT
    \STATE Draw $b$ ID samples from $\mathcal{D}_I$
    \STATE Draw $b \cdot K$ low-entropy samples from $\text{LDS}(q_{\theta}(y|\mathbf{x}))$
    \STATE Estimate objective function: $\mathcal{L}(\theta)$
    \STATE Obtain gradients: $\nabla_{\theta} \mathcal{L}(\theta)$
    \STATE Update parameters: $\theta  = \theta + \mu \nabla_{\theta} \mathcal{L}(\theta)$
    \UNTIL{convergence}
    \STATE {\bfseries Output:} fine-tuned discriminator $q_{\theta}(y|\mathbf{x})$
\end{algorithmic}
\end{algorithm}

\section{Experiments}\label{sec:experiment}
In this section, we demonstrate the effectiveness of FIG\footnote{The source codes are available at: \url{https://github.com/Lawliet-zzl/FIG}.} in comparison with the existing methods in detecting OOD samples. Furthermore, we analyze the sensitivity of hyper-parameters in the LDS and the objective function of FIG. Also, we analyze the transferability of the generated low-entropy samples, i.e., the low-entropy samples drawn from an implicit generator of a discriminator cannot be applied to patch the vulnerability of other discriminators with different network architectures. Finally, we present the visualization results to confirm that the generated low-entropy samples can effectively train OOD-sensitive discriminators.

\subsection{Setup}
The ID datasets for pretraining and fine-tuning discriminators are SVHN~\cite{SVHN:11}, CIFAR10~\cite{CIFAR10:09}, CIFAR100~\cite{CIFAR10:09}, and MiniImageNet~\cite{OSL:16}. The number of classes in these four datasets are 10, 10, 100, and 100, respectively. We follow the standard data augmentation practice for training samples. Specifically, we apply Resize(256) and RandomCrop((224,224)) to the samples in MiniImageNet and RandomCrop(32, padding=4) and RandomHorizontalFlip() to the samples in the other three datasets. To test OOD detection performance, the corresponding test samples of an ID training dataset are treated as ID, and the samples from the other real image datasets are treated as OOD. For SVHN, CIFAR10, and CIFAR100, the OOD datasets used are LSUN~\cite{LSUN:15}, TinyImageNet~\cite{IMAGENET:09}, Caltech256~\cite{CAL:06}, and COCO~\cite{COCO:14}. For MiniImageNet, the adopted OOD datasets include Oxfordflowers102~\cite{Oxfordflowers102}, Caltech256~\cite{CAL:06}, and DTD47~\cite{DTD47}. Because OOD samples come from distinct datasets with varying input sizes, by following the methods in ODIN~\cite{ODIN:18}, we resize or crop each OOD sample to maintain the same size as the ID samples. (r) and (c) represent resized and randomly cropped samples, respectively. For a fair comparison, following the setup of the baseline and state-of-the-art methods~\cite{BL:17,ODIN:18,DCC:20,DFR:20,GO:18,GM:20}, validation datasets are unavailable to validate the hyper-parameters because OOD detection should consider the detection performance on diverse OOD samples that are unobservable in the validation phase.

Four advanced neural network architectures, namely ResNet18~\cite{RES:16}, VGG19~\cite{VGG:15}, ShuffleNetV2~\cite{SHU:18}, and DenseNet100~\cite{DEN:17}, are used to create the discriminators. In pretraining a standard discriminator, its learning rate starts at $0.1$ and is divided by $10$ after $100$ and $150$ epochs. All networks are trained for $200$ epochs on the training sets with $128$ samples per mini-batch.

If not specified, the FIG setup is as follows. The same ID dataset is used to train and fine-tune standard discriminators. The fine-tuning process uses the learning rate $\mu = 0.001$, which is equal to the final learning rate in the pretraining phase. For the modified energy function Eq. (\ref{eq:mef}), we set the constant $c = 5$ because this value is sufficient to ensure that the exponential value is within the computer numerical range. For the LDS, following the suggestions of Welling and Teh~\cite{SGLD:11}, we set the step-size initialization $\epsilon_0 = 0.1$, the decay rate $\gamma = 0.9$, and the decay period $L = 100$. We set the low-entropy percentage $K = 0.1$ to balance the effectiveness and efficiency according to prior knowledge. We further discuss the effect of $K$ in Section~\ref{sec:para}. Following the state-of-the-art methods~\cite{GO:18,GM:20}, the baseline detector is applied to FIG to calculate the OOD scores for the test samples.

The OOD detection methods can be divided into two categories: OOD detectors and confidence enhancement methods. OOD detectors aim to extract OOD-sensitive information from trained discriminators. Accordingly, they can be applied to trained discriminators. Diverse OOD detectors are incorporated into the proposed FIG method to verify that FIG can adapt different OOD detectors. FIG is a confidence enhancement method, which aims to improve the OOD sensitivity by retraining or finetuning discriminators. Accordingly, the state-of-the-art confidence enhancement methods are compared with FIG to verify its effectiveness.

\subsection{Evaluation Metrics}
An OOD detector provides an OOD score for a test sample. ID and OOD samples are expected to have high and low scores, respectively. To evaluate the detection performance of OOD samples, we adopt the area under the receiver operating characteristic curve (AUROC)~\cite{BL:17,ODIN:18,MS:18}. A larger AUROC value indicates better OOD detection performance.

Since retraining or fine-tuning discriminators may change the classification accuracy, we also evaluate the harmonic mean of AUROC and accuracy to verify the comprehensive performance of classifying ID samples and detecting OOD samples. Furthermore, we adopt throughput~\cite{TP:21}, which represents the number of processed samples in one second, to measure the efficiency of generating low-entropy samples.

\begin{table}[t] \tiny
  \renewcommand{\arraystretch}{1.3}
  \setlength\tabcolsep{4pt}
  \caption{OOD detection performance of standard and fine-tuned discriminators with diverse detectors. Each value represents the average AUROC across eight OOD datasets, including LSUN(r), LSUN(c), TinyImageNet(r), TinyImageNet(c), Caltech256(r), Caltech256(c), COCO(r), and COCO(c). All the values are in percentage, and the boldface values represent relatively better detection performance.}
  \centering
  \label{tb:CompB}
\begin{tabular}{ccccccc}
\hline
\multirow{2}{*}{in-dist} & \multirow{2}{*}{network} & Baseline             & ODIN                 & MLB                  & DRF                  & GM                 \\ \cline{3-7}
                                 &                          & \multicolumn{5}{c}{Standard / Fine-tuned (FIG)}                                                     \\ \hline
\multirow{4}{*}{SVHN}            & ResNet18                 & 92.1 / \textbf{98.5} & 93.9 / \textbf{98.8} & 94.8 / \textbf{97.5} & 93.8 / \textbf{98.5} & 87.6 / \textbf{98.6} \\ \cline{2-7}
                                 & VGG19                    & 92.0 / \textbf{98.1} & 92.8 / \textbf{98.4} & 92.6 / \textbf{97.2} & 92.8 / \textbf{98.2} & 91.3 / \textbf{98.5} \\ \cline{2-7}
                                 & ShuffleNetV2             & 96.7 / \textbf{98.8} & 98.1 / \textbf{99.3} & 93.1 / \textbf{96.4} & 97.0 / \textbf{98.8} & 98.2 / \textbf{99.4} \\ \cline{2-7}
                                 & DenseNet100              & 91.3 / \textbf{97.6} & 92.8 / \textbf{97.7} & 95.4 / \textbf{96.8} & 91.4 / \textbf{97.4} & 77.9 / \textbf{96.2} \\ \hline
\multirow{4}{*}{CIFAR10}         & ResNet18                 & 91.2 / \textbf{95.0} & 92.3 / \textbf{95.8} & 91.5 / \textbf{94.3} & 91.7 / \textbf{95.0} & 91.0 / \textbf{95.6} \\ \cline{2-7}
                                 & VGG19                    & 88.2 / \textbf{92.2} & 89.0 / \textbf{92.8} & 88.5 / \textbf{90.5} & 89.0 / \textbf{92.9} & 89.0 / \textbf{92.3} \\ \cline{2-7}
                                 & ShuffleNetV2             & 88.7 / \textbf{92.1} & 91.4 / \textbf{95.2} & 89.5 / \textbf{92.7} & 87.2 / \textbf{91.4} & 90.4 / \textbf{94.7} \\ \cline{2-7}
                                 & DenseNet100              & 90.8 / \textbf{94.9} & 90.0 / \textbf{94.2} & 91.4 / \textbf{93.9} & 91.5 / \textbf{93.2} & 90.9 / \textbf{94.8} \\ \hline
\multirow{4}{*}{CIFAR100}        & ResNet18                 & 82.6 / \textbf{89.3} & 84.6 / \textbf{90.2} & 80.5 / \textbf{87.7} & 71.2 / \textbf{74.4} & 80.4 / \textbf{87.0} \\ \cline{2-7}
                                 & VGG19                    & 76.1 / \textbf{82.7} & 78.9 / \textbf{84.6} & 77.4 / \textbf{84.3} & 78.4 / \textbf{82.8} & 72.3 / \textbf{81.3} \\ \cline{2-7}
                                 & ShuffleNetV2             & 74.4 / \textbf{81.4} & 83.2 / \textbf{86.2} & 79.7 / \textbf{82.8} & 80.8 / \textbf{86.4} & 81.4 / \textbf{90.2} \\ \cline{2-7}
                                 & DenseNet100              & 83.0 / \textbf{93.1} & 86.3 / \textbf{93.4} & 83.5 / \textbf{93.9} & 75.0 / \textbf{78.1} & 72.7 / \textbf{85.5} \\ \hline
\end{tabular}
\end{table}

\subsection{Incorporating OOD detectors into FIG}
We incorporate diverse state-of-the-art OOD detectors into a standard discriminator and its fine-tuned discriminator. The standard discriminator is learned from a training ID dataset, and its corresponding fine-tuned discriminator is obtained by fine-tuning the discriminator with the OOD samples generated by its implicit generator.

We incorporate five different OOD detectors, the baseline~\cite{BL:17}, ODIN~\cite{ODIN:18}, MLB~\cite{MLB:18}, DRF~\cite{DFR:20}, and GM~\cite{GM:20}, into FIG. The baseline~\cite{BL:17} directly defines the maximum softmax output value from a discriminator as the OOD score without any hyper-parameters. For ODIN~\cite{ODIN:18}, we select the temperature in $\{1,2,5,10,20,50,100,200,500,100\}$ and the perturbation magnitude of 21 evenly spaced numbers starting from $0$ and ending at $0.004$. The best results are reported. For MLB~\cite{MLB:18}, we tune the magnitude of noise in $\{0, 0.0005, 0.001, 0.0014, 0.002, 0.0024, 0.005, 0.01, 0.05, 0.1, \\  0.2\}$. For a fair comparison, we add the scores from different layers without training a logistic regression on a validation OOD dataset in MLB. For DRF~\cite{DFR:20}, the magnitude of noise is $0.05$ for CIFAR10 and SVHN and $0.0025$ for CIFAR100. For GM~\cite{GM:20}, the order of computing feature correlations falls in the set $\{1,\dots,10\}$.

We summarize the results in Table~\ref{tb:CompB}. It shows that a fine-tuned discriminator achieves a significant improvement ($1.22\%$ to $23.49\%$) over its corresponding standard discriminator. Specifically, the discriminator fine-tuned by FIG achieves significant detection performance for both the detectors that apply the softmax outputs and the feature embeddings from network layers. This shows that FIG can improve the OOD sensitivity of a standard discriminator and alleviate the feature collapse problem~\cite{FC:21}. According to the learning principle of FIG, the fundamental reason for its OOD detection improvement is that the distributional vulnerability of a standard discriminator has been effectively patched by the samples generated by its corresponding implicit generator.

Although FIG can achieve measurable improvement over all the considered settings, its performance under different scenarios is distinct. For example, FIG approaches near-optimal performance in SVHN-ResNet18 and still causes a large gap from the optimal performance in CIFAR100-VGG19. This is because the performance of FIG depends on the OOD sensitivity of the standard discriminator. Specifically, FIG infers the implicit generator of a standard discriminator to improve its OOD sensitivity. The OOD sensitivities of discriminators depend on the network architectures and the data characteristics of the training ID dataset. Therefore, one solution to further improve FIG performance is to retrain the standard discriminator with extra OOD knowledge and incorporate FIG into the retraining procedure, which is our future work.

\begin{table*}[t] \tiny
  \renewcommand{\arraystretch}{1.5}
  \caption{OOD detection performance for networks learned on SVHN, CIFAR10, and CIFAR100. The value for an OOD dataset indicates its corresponding AUROC presented as a percentage, and the values for Ave. indicate the average AUROC across all the test OOD datasets. Boldface values represent the relatively better detection performance.}
  \centering
  \label{tb:compR}
\begin{tabular}{cccccc}
\hline
\multirow{2}{*}{In-dist} & \multirow{2}{*}{Out-of-dist} & \multicolumn{4}{c}{GS / MIXUP / AD / JCL / DCC* / FIG} \\ \cline{3-6}
                          &                 & ResNet18 & VGG19 & ShuffleNetV2 & DenseNet100 \\ \hline
\multirow{9}{*}{SVHN}     & LSUN(r)         & 99.2 / 95.4 / 94.7 / 98.7 / 94.5 / \textbf{99.4} & 98.5 / 96.3 / \textbf{99.7} / 98.3 / 91.0 / 99.3 & 98.0 / 95.9 / 96.4 / 91.6 / 99.1 / \textbf{99.7} & 95.4 / 95.9 / 90.0 / 91.4 / \textbf{98.4} / \textbf{98.4}\\
                          & LSUN(c)         & 98.4 / 92.7 / 95.8 / \textbf{99.7} / 97.6 / 97.1 & 98.9 / 94.3 / 96.4 / \textbf{98.2} / 97.0 / 98.0 & 96.7 / 92.7 / 95.2 / 94.3 / \textbf{98.8} / 97.7 & 92.9 / 95.2 / 90.8 / 91.8 / \textbf{99.1} / 95.9\\
                          & TinyImageNet(r) & 99.0 / 95.2 / 95.1 / 98.9 / 94.9 / \textbf{99.4} & 98.6 / 96.8 / \textbf{99.7} / 98.3 / 93.4 / 99.1 & 98.0 / 96.2 / 96.8 / 92.9 / \textbf{99.9} / 99.6 & 95.5 / 95.7 / 91.2 / 90.3 / 98.0 / \textbf{98.8}\\
                          & TinyImageNet(c) & 98.7 / 94.8 / 96.6 / \textbf{99.7} / 97.4 / 98.2 & 99.2 / 95.7 / \textbf{99.5} / 98.4 / 95.7 / 98.8 & 97.3 / 96.0 / 96.1 / 95.4 / 98.4 / \textbf{99.0} & 93.5 / 95.9 / 91.3 / 91.9 / \textbf{99.5} / 97.5\\
                          & Caltech256(r)  & 95.9 / 90.9 / 93.3 / 90.4 / 92.5 / \textbf{97.2} & 95.4 / 92.9 / 93.8 / 95.2 / 91.5 / \textbf{95.5} & 95.2 / 92.7 / 94.2 / 91.2 / \textbf{97.8} / 97.3 & 91.9 / 93.5 / 88.8 / 89.6 / \textbf{95.5} / 95.3\\
                          & Caltech256(c)  & 97.7 / 91.7 / 94.1 / 97.3 / 94.5 / \textbf{98.8} & 96.9 / 94.1 / 92.4 / 97.7 / 88.7 / \textbf{98.3} & 96.5 / 90.5 / 92.5 / 92.6 / 98.5 / \textbf{98.7} & 94.8 / 94.7 / 89.5 / 90.8 / 97.0 / \textbf{99.2}\\
                          & COCO(r)        & 97.5 / 92.9 / 94.6 / 94.6 / 95.1 / \textbf{98.4} & 96.6 / 94.9 / \textbf{99.5} / 96.3 / 91.6 / 97.2 & 96.7 / 94.6 / 96.3 / 91.5 / 97.4 / \textbf{98.8} & 94.2 / 95.3 / 88.4 / 90.0 / 96.1 / \textbf{96.7}\\
                          & COCO(c)        & 97.9 / 91.1 / 94.2 / 97.6 / 93.6 / \textbf{99.1} & 97.2 / 93.7 / 97.2 / 98.1 / 88.4 / \textbf{98.4} & 96.6 / 90.5 / 95.2 / 92.6 / 98.7 / \textbf{99.2} & 95.0 / 93.9 / 91.0 / 90.9 / 97.2 / \textbf{99.4}\\ \cline{2-6}
                          & Ave.            & 98.0 / 93.1 / 94.8 / 97.1 / 95.0 / \textbf{98.5} & 97.7 / 94.8 / 97.3 / 97.6 / 92.2 / \textbf{98.1} & 96.9 / 93.7 / 95.3 / 92.8 / 98.6 / \textbf{98.8} & 94.2 / 95.0 / 90.1 / 90.8 / \textbf{97.6} / \textbf{97.6}\\ \hline
\multirow{9}{*}{CIFAR10}  & LSUN(r)         & 92.8 / 92.8 / 91.9 / 90.8 / 98.7 / \textbf{99.0} & 89.4 / 95.3 / 80.4 / 90.8 / 96.4 / \textbf{97.4} & 83.0 / 83.5 / 81.4 / 88.8 / 98.6 / \textbf{99.8} & 92.2 / 87.8 / 90.6 / 94.7 / \textbf{99.4} / 99.1\\
                          & LSUN(c)         & 95.0 / 95.7 / 94.1 / 90.8 / 98.2 / \textbf{98.9} & 92.3 / 95.7 / 86.4 / 90.1 / \textbf{97.3} / 96.7 & 89.0 / 86.7 / 82.5 / 91.9 / \textbf{98.0} / 93.0 & 93.1 / 96.1 / 91.5 / 97.3 / \textbf{98.3} / 98.0\\
                          & TinyImageNet(r) & 91.9 / 89.8 / 89.1 / 92.7 / 95.4 / \textbf{99.0} & 86.8 / 93.9 / 78.7 / 84.3 / 92.4 / \textbf{96.4} & 82.0 / 82.6 / 77.2 / 84.7 / \textbf{97.3} / 96.2 & 91.5 / 87.6 / 85.9 / 93.6 / \textbf{99.1} / 97.3\\
                          & TinyImageNet(c) & 93.2 / 93.4 / 93.0 / 92.7 / \textbf{96.2} / 95.7 & 89.7 / 94.3 / 84.4 / 92.7 / 91.3 / \textbf{94.9} & 87.4 / 85.9 / 85.8 / 88.2 / \textbf{96.5} / 92.2 & 92.3 / 93.4 / 89.3 / 96.2 / \textbf{98.7} / 96.5\\
                          & Caltech256(r)  & 86.9 / 80.0 / 85.9 / \textbf{92.9} / 85.0 / 88.0 & 82.5 / \textbf{86.1} / 76.1 / 84.3 / 80.4 / 83.4 & 79.3 / 78.9 / 76.3 / 81.2 / \textbf{84.6} / 83.0 & 86.7 / 79.5 / 85.1 / \textbf{90.1} / 87.6 / 87.8\\
                          & Caltech256(c)  & 93.0 / 90.3 / 91.5 / 84.3 / 91.7 / \textbf{94.7} & 88.5 / \textbf{92.7} / 79.4 / 89.5 / 87.6 / 90.7 & 82.5 / 80.4 / 78.1 / 79.1 / 87.1 / \textbf{91.9} & 91.0 / 89.9 / 90.8 / \textbf{95.2} / 91.3 / 94.4\\
                          & COCO(r)        & 87.9 / 83.9 / 87.2 / \textbf{91.7} / 85.9 / 90.5 & 85.0 / \textbf{88.2} / 79.4 / 85.2 / 81.0 / 86.5 & 80.5 / 79.9 / 80.8 / 82.3 / 85.1 / \textbf{88.3} & 87.6 / 83.8 / 85.8 / 88.5 / 88.8 / \textbf{89.6}\\
                          & COCO(c)        & 92.7 / 87.5 / 91.6 / 85.2 / 89.9 / \textbf{94.5} & 88.4 / 93.8 / 79.2 / 90.8 / 87.3 / \textbf{91.7} & 84.1 / 81.6 / 78.7 / 79.6 / 87.9 / \textbf{92.4} & 91.0 / 89.5 / 90.7 / 93.9 / 90.6 / \textbf{96.2}\\ \cline{2-6}
                          & Ave.            & 91.7 / 89.2 / 90.5 / 90.1 / 92.6 / \textbf{95.0} & 87.8 / 92.5 / 80.5 / 88.5 / 89.2 / \textbf{92.2} & 83.5 / 82.4 / 80.1 / 84.5 / 91.9 / \textbf{92.1} & 90.7 / 88.4 / 88.7 / 93.7 / 94.2 / \textbf{94.9}\\ \hline
\multirow{9}{*}{CIFAR100} & LSUN(r)         & 83.6 / 78.0 / 82.7 / 87.6 / 93.4 / \textbf{93.8} & 79.2 / 75.4 / 71.5 / 80.7 / \textbf{87.3} / 82.5 & 71.9 / 55.9 / 68.8 / 65.7 / 80.4 / \textbf{82.3} & 81.9 / 75.2 / 82.6 / 86.1 / \textbf{98.7} / 98.6\\
                          & LSUN(c)         & 85.4 / 77.6 / 81.5 / 80.5 / \textbf{88.3} / 85.0 & 83.7 / 80.9 / 78.3 / 81.9 / 85.6 / \textbf{85.9} & 75.1 / 71.2 / 76.7 / 77.3 / \textbf{87.7} / 82.9 & 81.6 / 81.9 / 81.4 / 88.4 / \textbf{95.3} / 94.6\\
                          & TinyImageNet(r) & 82.9 / 74.4 / 81.5 / 87.2 / 92.8 / \textbf{97.1} & 76.6 / 75.6 / 70.9 / 80.5 / \textbf{81.6} / 80.0 & 72.5 / 61.1 / 64.4 / 63.5 / 78.4 / \textbf{84.7} & 82.5 / 74.1 / 82.3 / 83.5 / 98.6 / \textbf{98.7}\\
                          & TinyImageNet(c) & 87.1 / 83.7 / 83.8 / 83.3 / \textbf{91.4} / 89.6 & 83.4 / 81.8 / 77.3 / 79.9 / 83.9 / \textbf{87.2} & 78.9 / 78.8 / 78.5 / 75.9 / \textbf{88.5} / 86.4 & 84.1 / 84.9 / 84.0 / 87.8 / \textbf{97.6} / 96.6\\
                          & Caltech256(r)  & 75.3 / 75.2 / 76.2 / 79.7 / \textbf{83.3} / 82.4 & 71.5 / 71.2 / 69.1 / \textbf{87.8} / 77.3 / 76.9 & 67.2 / 68.9 / 68.5 / 67.8 / \textbf{74.6} / 72.5 & 74.4 / 72.3 / 75.8 / 81.3 / 82.9 / \textbf{83.4}\\
                          & Caltech256(c)  & 82.1 / 83.7 / 81.6 / 83.6 / 87.9 / \textbf{89.2} & 79.7 / 79.9 / 74.7 / 80.6 / 76.7 / \textbf{84.5} & 71.7 / 70.0 / 71.5 / 71.2 / 76.3 / \textbf{81.5} & 81.6 / 80.1 / 81.3 / 85.7 / 86.9 / \textbf{92.9}\\
                          & COCO(r)        & 77.4 / 78.8 / 78.8 / 80.2 / 83.2 / \textbf{84.1} & 75.6 / 77.7 / 72.9 / 76.8 / \textbf{79.7} / 79.2 & 70.8 / 69.7 / 71.5 / 69.2 / \textbf{77.8} / 75.6 & 77.0 / 79.8 / 78.5 / 80.6 / 84.5 / \textbf{85.4}\\
                          & COCO(c)        & 83.2 / 80.5 / 82.2 / 82.5 / 89.0 / \textbf{93.4} & 81.7 / 79.3 / 75.8 / 81.7 / 78.4 / \textbf{85.6} & 71.8 / 70.5 / 71.1 / 71.6 / 78.3 / \textbf{85.6} & 82.0 / 79.6 / 82.3 / 84.5 / 88.1 / \textbf{94.7}\\ \cline{2-6}
                          & Ave.            & 82.1 / 79.0 / 81.0 / 83.1 / 88.7 / \textbf{89.3} & 78.9 / 77.7 / 73.8 / 81.2 / 81.3 / \textbf{82.7} & 72.5 / 68.3 / 71.4 / 70.3 / 80.2 / \textbf{81.4} & 80.6 / 78.5 / 81.0 / 84.7 / 91.6 / \textbf{93.1}\\ \hline
\end{tabular}
\end{table*}

\subsection{Comparison Results}
To verify the quality of the OOD samples generated by the implicit generators, we compare FIG with five state-of-the-art confidence enhancement methods that retrain or fine-tune standard discriminators, namely Gaussian (GS)~\cite{DA:19}, MIXUP~\cite{MIXUP:18}, adversarial (AD)~\cite{AD:15}, joint confidence loss (JCL)~\cite{GO:18}, and DeConf-C* (DCC*)~\cite{DCC:20}. For a fair comparison, following the setup of the state-of-the-art methods~\cite{GO:18,GM:20}, we apply the embedded detector based on ODIN~\cite{ODIN:18} for DCC* and the baseline detector~\cite{BL:17} for the other compared methods to calculate the OOD scores without loss of generality.

The settings of all the compared methods are the same as their original. To use samples drawn from the Gaussian distribution as OOD samples in GS, we adopt Algorithm~\ref{alg:FIG} to fine-tune the standard discriminators for a fair comparison. Specifically, we replace the low-entropy samples drawn from the LDS with Gaussian noise samples in Algorithm~\ref{alg:FIG}. As for MIXUP, the mixing coefficients that control the interpolation strength between sample pairs are drawn from Beta$(1,1)$ for all ID datasets. When using adversarial samples as the generated OOD samples in AD to retrain the standard discriminators, we set the perturbation magnitude as $0.1$ and the weights of both the cross-entropy loss and the adversarial objective function as $0.5$. Another advanced method JCL retrains a standard discriminator with a generative adversarial network (GAN)~\cite{GAN:14} and encourages the softmax probabilities of generated samples to satisfy a uniform distribution. For JCL, we use mini-batch size $128$ and regularization coefficient $1$ of the Kullback-Leibler (KL) divergence term for SVHN. The two hyper-parameters are $64$ and $0.1$ respectively for the other three training ID datasets. For DCC*, we adopt the cosine similarity in the scoring function and search for the adversarial perturbation magnitude with only ID samples.

The OOD detection results on SVHN, CIFAR10, and CIFAR100 are displayed in Table~\ref{tb:compR}. Comparing all the methods, we observe that FIG does not achieve the best OOD detection performance on some ID and OOD dataset pairs. Lee et al.~\cite{GO:18} offer a possible explanation, i.e., the distribution of a specific OOD dataset does not effectively cover all tested out-of-distributions. We thus verify the effect of FIG on different test OOD datasets, and FIG inevitably reduces the effect on some OOD samples in order to pursue the overall OOD detection improvement. Compared with GS, FIG obtains significant improvement ($5.69\%$). We thus verify that the generated samples from the implicit generators are not simple high-confidence noise but informative images that can reveal the discriminator vulnerability. For all neural architectures, compared with the other state-of-the-art methods, FIG achieves the best OOD detection performance. It makes an average of $3.29\%$, $5.34\%$, and $9.01\%$ AUROC improvement on the three training ID datasets, SVHN, CIFAR10, and CIFAR100, respectively.

We also perform experiments on a larger resolution dataset MiniImageNet, and the results are presented in Table~\ref{tb:comp224}. FIG achieves the most significant average AUROC value across all the test OOD datasets with an average of $7.87\%$ AUROC improvement over the other state-of-the-art methods. As a result, FIG achieves the best OOD detection performance. This is because the generated low-entropy samples of FIG are specific to the ID training dataset and network architecture. The data characteristics indicate that the generated low-entropy samples can be applied to patch the vulnerability of a standard network to improve OOD sensitivity. Furthermore, the quantitative analysis in Table~\ref{tb:time} shows that FIG can efficiently generate low-entropy samples and larger resolutions do not significantly improve the running time of generating low-entropy samples. This is due to the early stop strategy in Eq. (\ref{eq:sample}). Therefore, FIG is applicable for high-resolution samples.

The harmonic means of AUROC and accuracy of the compared methods are shown in Table~\ref{tb:CompACC}. JCL and DCC* achieve the significant performance in detecting some OOD samples, as shown in Table~\ref{tb:compR}. However, the corresponding harmonic means are close to the baseline method which only applies a standard discriminator without modification. The results indicate that the two methods significantly sacrifice the classification ability to improve OOD sensitivity. However, FIG achieves the most significant harmonic means on all ID training datasets, which indicates that FIG finds the best balance between classifying ID samples and detecting OOD samples. The reasons are two-fold: (1) low-entropy samples are generated from the implicit generator for a given standard discriminator; (2) fine-tuning the standard discriminator with the specific generated samples will not seriously disturb the learning process of classifying ID samples.

In general, our FIG can improve the OOD detection performance and maintain high ID classification accuracy. We recall the diverse vulnerability of discriminators with different architectures to understand the reason behind this. Hence, low-entropy samples generated by particular generators cannot correspondingly address the discriminator-specific vulnerability. FIG patches the vulnerability of a standard discriminator to improve its OOD detection performance with the samples generated by its implicit generator, and the implicit generator knows what kind of samples are OOD for the discriminator. These conclusions also explain why FIG can balance OOD detection and ID classification after being fine-tuned on the generated low-entropy samples. The low-entropy samples are data- and network-adaptive, which enables the standard discriminator to learn the knowledge from the ID samples with less interference.


\begin{table}[h] \scriptsize
  \renewcommand{\arraystretch}{1.5}
  \caption{OOD detection performance for ResNet18 learned from MiniImageNet. The value for an OOD dataset indicates its corresponding AUROC presented as a percentage, and the values for ``Ave.'' indicate the average AUROC across all the test OOD datasets. Boldface values represent the relatively better detection performance.}
  \centering
  \label{tb:comp224}
\begin{tabular}{ccccccc}
\hline
Out-of-dist    & GS   & MIXUP & AD   & JCL  & DCC*  & FIG  \\ \hline
Oxfordflowers102(r)     & 80.0 & 79.1  & 77.2 & 77.3 & 80.7 & \textbf{81.1} \\ \hline
Oxfordflowers102(c)     & 82.6 & 81.7  & 78.8 & 83.6 & 84.2 & \textbf{85.5} \\ \hline
Caltech256(r)           & 77.9 & 78.7  & 63.5 & 75.1 & 83.2 & \textbf{84.6} \\ \hline
Caltech256(c)           & 80.9 & 79.2  & 82.1 & 84.2 & 85.0 & \textbf{89.8} \\ \hline
DTD47(r)                & 73.5 & 75.2  & 69.0 & 69.6 & 74.4 & \textbf{81.2} \\ \hline
DTD47(c)                & 78.7 & 78.9  & 73.5 & 81.2 & 80.7 & \textbf{84.0} \\ \hline
Ave.                    & 78.9 & 78.8 & 74.0 & 78.5 & 81.4 & \textbf{84.4} \\ \hline
\end{tabular}
\end{table}

\begin{table}[h] \tiny
  \renewcommand{\arraystretch}{1.5}
  \caption{Harmonic means of AUROC and accuracy. Boldface values represent the relatively better balance between classifying ID samples and detecting OOD samples.}
  \centering
  \label{tb:CompACC}
\begin{tabular}{cccccccc}
\hline
In-dist      & Standard & GS   & MIXUP & AD   & JCL  & DCC*  & FIG  \\ \hline
SVHN         & 47.1     & 48.5 & 47.4  & 47.7 & 48.0 & 47.7 & \textbf{48.7} \\ \hline
CIFAR10      & 46.6     & 46.7 & 46.2  & 44.9 & 45.8 & 46.8 & \textbf{47.5} \\ \hline
CIFAR100     & 39.9     & 39.7 & 39.4  & 35.9 & 38.9 & 40.5 & \textbf{41.2} \\ \hline
MiniImageNet & 38.8     & 38.6 & 39.0  & 36.5 & 37.3 & 39.3 & \textbf{40.0} \\ \hline
\end{tabular}
\end{table}

\begin{table}[]
  \renewcommand{\arraystretch}{1.3}
  \caption{Efficiency of generating low-entropy samples on ResNet18 in terms of throughput.}
  \centering
  \label{tb:time}
\begin{tabular}{ccc}
\hline
In-dist      & \begin{tabular}[c]{@{}c@{}}image \\ size\end{tabular} & \begin{tabular}[c]{@{}c@{}}throughput\\ (image/s)\end{tabular} \\ \hline
CIFAR10      & $32 \times 32$                                                    & 556                                                         \\ \hline
MiniImageNet & $224 \times 224$                                                  & 222                                                         \\ \hline
\end{tabular}
\end{table}

\subsection{Hyper-parameter Analysis}~\label{sec:para}
This section empirically shows the impact of the low-entropy percentage $K$ on the proposed FIG method. We test the effect of $K$ by setting it to $0, 0.01, 0.1, 0.4, 0.7, 1$ respectively. We show the widespread applicability and stability of the hyper-parameter $K$ on CIFAR10, SVHN, and CIFAR100 with the network architectures Resnet18, VGG19, ShuffleNetV2, and DenseNet100 in terms of AUROC. Note that when $K = 0$, FIG only applies training ID samples to fine-tune discriminators without generating low-entropy samples.

\begin{figure}
\centering
\includegraphics[width=0.3\textwidth]{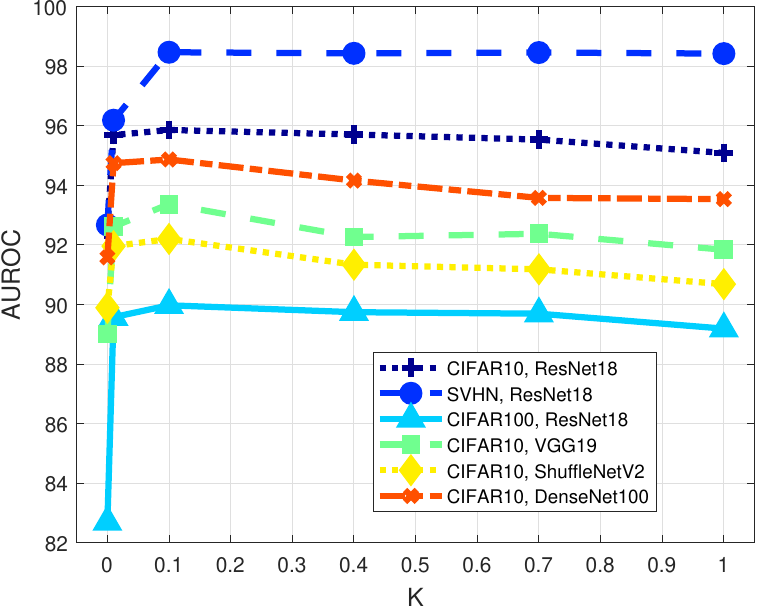}
  \caption{Effect of the OOD percentage $K$. Each point refers to an average AUROC score on the eight OOD datasets, namely LSUN(r), LSUN(c), TinyImageNet(r), TinyImageNet(c), Caltech256(r), Caltech256(c), COCO(r) and COCO(c).}
\label{fig:para}
\end{figure}

The results of verifying $K$ are shown in Fig.~\ref{fig:para}. We observe that increasing the low-entropy percentage $K$ can improve the detection performance. The detection performance diminishes when $K$ is sufficiently large ($K>0.1$). However, having a large $K$ with performance reduction is acceptable. Recall that an implicit generator depends on a standard discriminator, and the discriminator is updated by the low-entropy samples generated by the implicit generator. According to this design principle, implicit generators generate low-entropy samples, which could be ID or OOD samples. We assume that most drawn samples are OOD because ID samples have limited classes while the generated samples are diverse. When $K$ is sufficiently large ($K>0.1$), more generated ID samples are encouraged to yield low-confidence predictions in the fine-tuning phase. This result is contradictory to the expectation that a standard distribution should assign high-confidence predictions for training ID samples. Specifically, a large set of the generated samples contain more generated ID samples. This causes the bias in the estimated gradients of the entropy in the objective function. The dynamic implicit generator makes the biased estimation more serious. Therefore, a small low-entropy percentage, such as $K \in [0.01, 0.1]$, is a better choice for FIG. Hence, we apply $K=0.1$ to balance effectiveness and efficiency by default.

\begin{figure}
\centering
\includegraphics[width=0.42\textwidth]{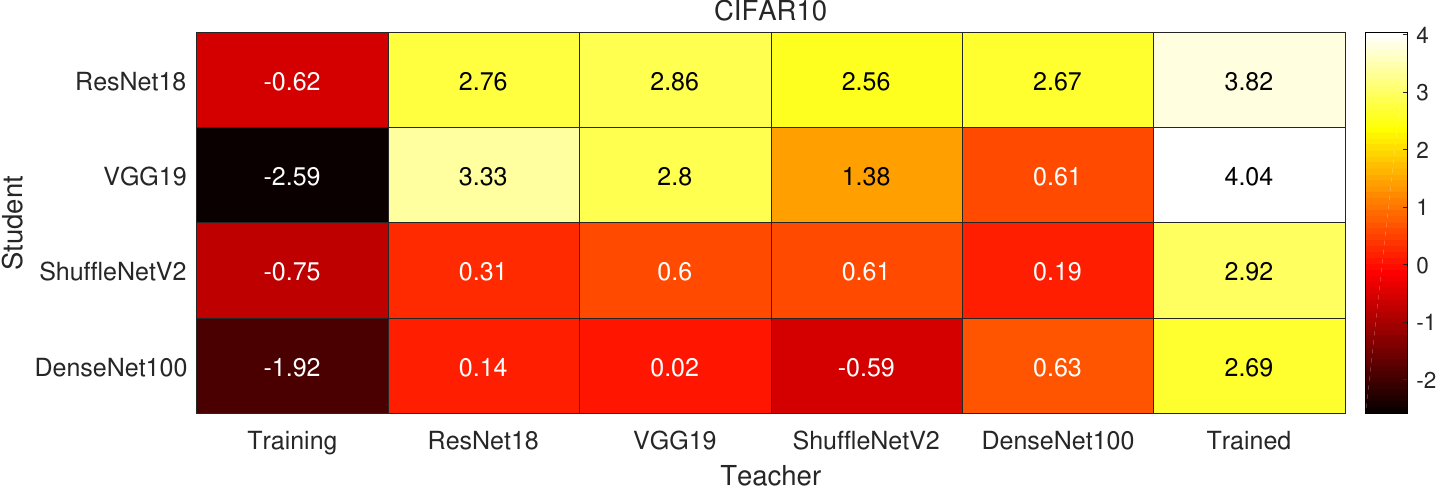}
\caption{Transferability of the generated OOD samples between two discriminators. Each student is the discriminator in the objective function, and each teacher is the discriminator used to infer the implicit generator in the Langevin dynamic sampler. The training and trained teachers are initialized randomly by standard discriminators. Both training and trained teachers are continuously updated as the standard discriminators change. The teachers named by network architectures are randomly initialized. Their parameters are fixed during the learning process of students. A value in the colored boxes represents the percentage of AUROC improvement over the standard discriminator with the same student network architecture. Lighter colors reflect better results. For all columns except the first and last, diagonal entries correspond to answer \textbf{A1}, and off-diagonal entries correspond to \textbf{A2}. The entries in the first column correspond to \textbf{A3}.}
\label{fig:trans}
\end{figure}

\subsection{Transferability Analyses}
In FIG, low-entropy samples are drawn from the implicit generator of a standard discriminator. The generated low-entropy samples are then used to patch the vulnerability of the discriminator. We analyze the transferability of the generated low-entropy samples to verify that (1) the implicit generator should be updated as the corresponding discriminator is updated; (2) low-entropy samples drawn from the implicit generator of a discriminator cannot be applied to patch the vulnerabilities of other discriminators with different network architectures; and (3) FIG is not suitable for randomly initialized discriminators.

The discriminator following LDS and the discriminator in the objective function Eq. (\ref{eq:ob1}) can be treated as a teacher and a student, respectively. Therefore, a discriminator trained on ID samples is a student learning without teachers, and a discriminator trained by FIG is a student learning with a teacher. The teacher teaches the student how to find the vulnerability. The student who receives the knowledge from the teacher then knows the previously unknown (i.e., the vulnerability). The teacher already has some knowledge of the network structure if the teacher is pretrained. The teacher and the student learn from each other as the discriminator used in LDS is updated. Accordingly, we analyze the teacher from different perspectives and ask the following three questions:
\begin{itemize}
  \item \textbf{Q1:} What if the teacher stops learning from the student? This corresponds to applying a fixed discriminator to infer an implicit generator in each iteration.
  \item \textbf{Q2:} What if the expertise of the teacher mismatches that of the student? This corresponds to generating low-entropy samples according to a discriminator to patch the vulnerability of other discriminators with different architectures.
  \item \textbf{Q3:} What if the teacher does not yet have enough knowledge or experience but still learns from the student? In this situation, the discriminator is trained from scratch, and the implicit generator is updated according to the training discriminator before each epoch.
\end{itemize}

To answer these questions, we design the following experiments, whose results are shown in Fig.~\ref{fig:trans}. We run FIG in terms of different teacher-student pairs on CIFAR10. We evaluate the improved performance over the baseline with the same network architecture as the student on detecting different OOD samples, namely LSUN(r), LSUN(c), TinyImageNet(r), TinyImageNet(c), Caltech256(r), Caltech256(c), COCO(r), and COCO(c). In summary, the following findings address the above questions.
\begin{itemize}
  \item \textbf{A1:} Similarly, we fix the discriminator in LDS and ensure this discriminator and the discriminator in the objective function have the same network structure. If standard discriminators are fixed in LDS for each iteration in the generation process, the detection performance will be worse than when the on-the-fly discriminators are used to infer implicit generators. The main reason for this is that vulnerability is dynamic as refining discriminators leads to new vulnerabilities. This dynamic property requires the implicit generators to be updated continuously.
  \item \textbf{A2:} We replace the regularly updated discriminator in the LDS input with a fixed discriminator, which is diversified with different architectures. These teachers have to be fixed because only the gradients of students are calculated in FIG and the gradients for teachers are not available. When students and teachers have different architectures, the OOD detection performance generally declines. This is because the generated low-entropy samples from a network do not match the vulnerabilities of networks with different architectures. Therefore, the generated low-entropy samples are model specific.
  \item \textbf{A3:} We replace the discriminator in the input list of Algorithm~\ref{alg:FIG} with a randomly initialized discriminator and use the same training setup as the baseline. The discriminator is trained for $200$ epochs. The learning rates start at $0.1$ and are divided by $10$ after $100$ and $150$ epochs. It is important to give knowledge to teachers as we find fine-tuning a standard discriminator can achieve better performance than retraining a new one. This is because the capable discriminators deduce reliable implicit generators, which guarantees the right direction to patch the vulnerability.
\end{itemize}
According to the transferability analyses, we apply FIG on a standard discriminator and continually update the standard discriminator and the corresponding implicit generator.

\subsection{Visualization of the Results}
The samples generated by implicit generators can be applied to train OOD-sensitive discriminators because these samples generated by standard discriminator have high confidence and are almost OOD. This is verified by visualizing the change in the confidence and energy along the fine-tuning process, the embedding results, and the content of generated low-entropy samples. The network architecture is Resnet18, and the training ID datasets are CIFAR10 and SVHN.

\begin{figure}
\centering
\subfigure{
    \begin{minipage}{0.47\linewidth}
    \centering
    \includegraphics[width=1\textwidth]{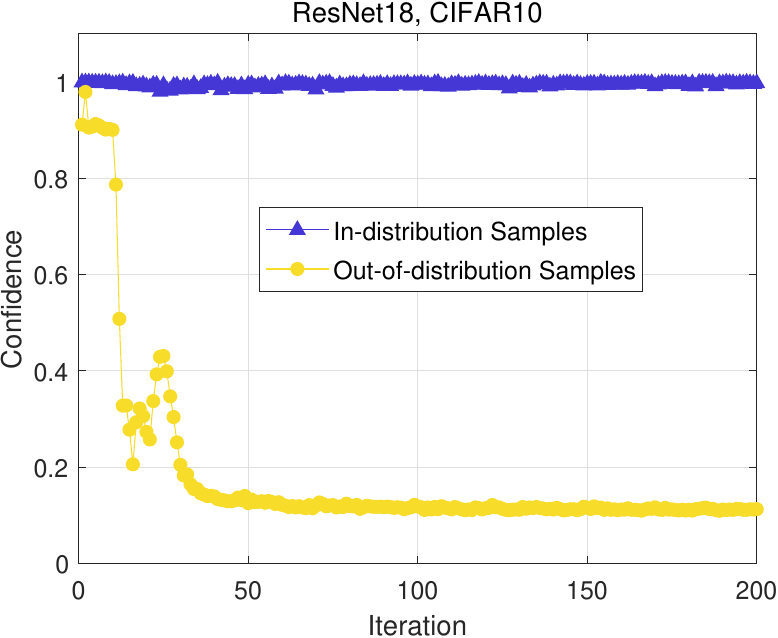}
    \vspace{0.05em}
    \end{minipage}
  }
\subfigure{
    \begin{minipage}{0.47\linewidth}
    \centering
    \includegraphics[width=1\textwidth]{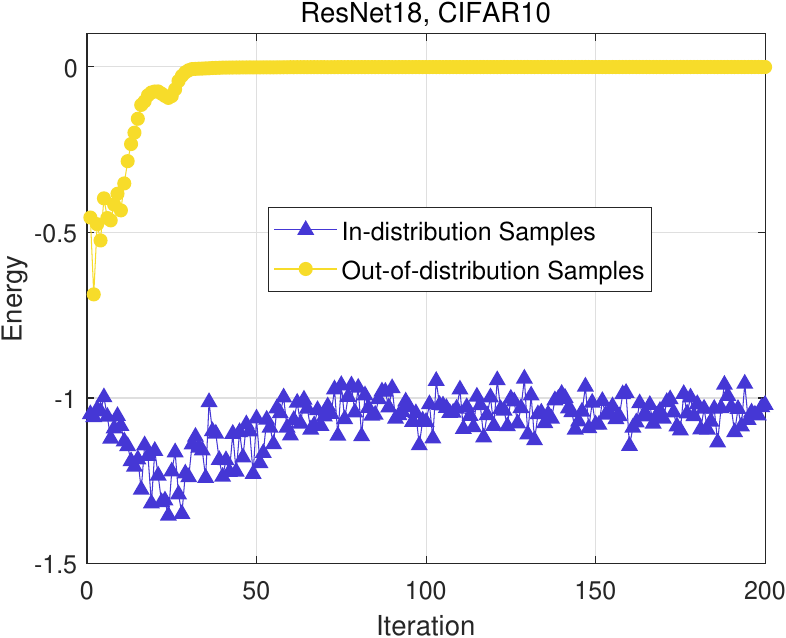}
    \vspace{0.05em}
    \end{minipage}
  }
\caption{Confidence and energy of training ID samples and generated samples. Each point indicates an average value of confidence or energy on the training ID dataset or a generated dataset.}
\label{fig:process}
\end{figure}

\subsubsection{Confidence and Energy}
We analyze FIG from confidence and energy perspectives, respectively. We visualize the changes in confidence and energy on both training ID samples and the generated samples along with the fine-tuning of the discriminators. For the confidence of the generated samples, FIG should encourage low scores since the OOD sensitivity of discriminators can be improved by making it difficult for the corresponding implicit generators to produce high-confidence OOD samples. For the energy of the generated samples, the implicit generators should have high values according to the design principle.

The results are reported in Fig.~\ref{fig:process}. We find that ID samples maintain high-confidence scores and stable energy values. For the generated samples, the confidence scores are close to one in the preliminary stage, which then drop continuously. It is increasingly difficult for implicit generators to generate high-confidence OOD samples since samples with a higher energy are explored as iterations increase. Although the energy of the generated samples is higher than that of the ID samples, the distribution of the prediction probability vectors approximates to a uniform distribution since the confidence scores are close to $0.1 = 1 / \text{class-number}$ on the training dataset CIFAR10. Therefore, we conclude that implicit generators can produce high-confidence OOD samples in the preliminary stage, which then fails after the vulnerability is patched.

\begin{figure}
  \centering
  \includegraphics[width=0.48\textwidth]{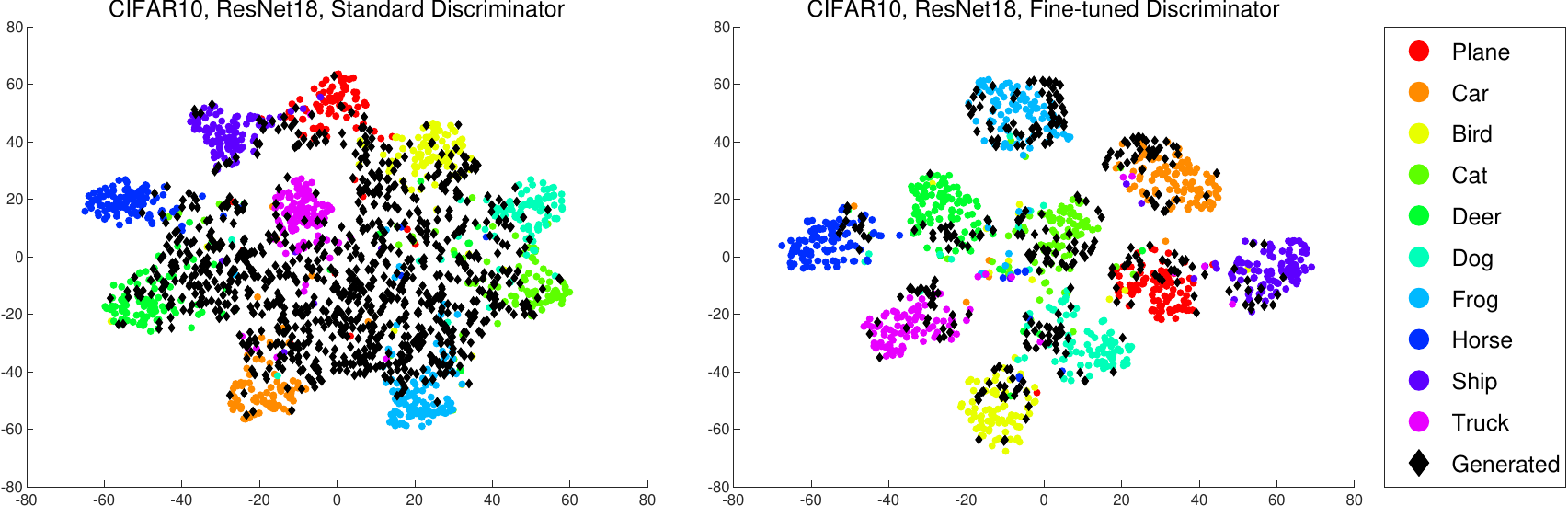}
\caption{Embedding results of test ID samples and the generated samples from a standard discriminator and a fine-tuned discriminator. The black diamonds indicate the generated samples, and the colored circles represent the test ID samples.}
\label{fig:embedding}
\end{figure}

\subsubsection{Embedding Visualization}~\label{sec:ev}
Fig.~\ref{fig:embedding} presents the embedding results of test ID samples and the generated samples of a standard discriminator and a fine-tuned discriminator by t-SNE~\cite{SNE:08}. We randomly sample $10 \%$ of the test ID samples and draw $1,000$ samples by the implicit generators. Only the samples with confidence scores over $0.9$ are plotted. The results show that the generated low-entropy samples by a standard discriminator and a fine-tuned discriminator are distinguished from and included in the ID classes, respectively. Therefore, their low-entropy samples tend to be OOD and ID, respectively. This is reasonable because the distributional vulnerability of the fine-tuned discriminator is patched by the generated low-entropy samples, which makes it harder to generate OOD samples. The embedding results for the fine-tuned discriminators verify that FIG effectively applies the generated samples to patch the vulnerability.

\begin{figure}
\centering
\subfigure[CIFAR10 (Dog)]{
    \begin{minipage}{0.47\linewidth}
    \centering
    \includegraphics[width=1\textwidth]{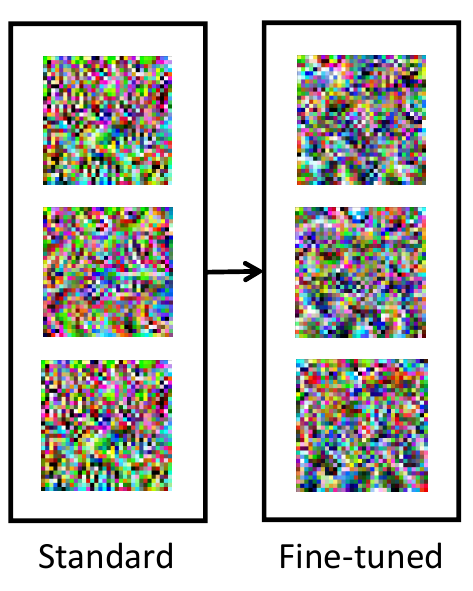}
    \vspace{0.05em}
    \end{minipage}
  }
\subfigure[SVHN (Digit 6)]{
    \begin{minipage}{0.47\linewidth}
    \centering
    \includegraphics[width=1\textwidth]{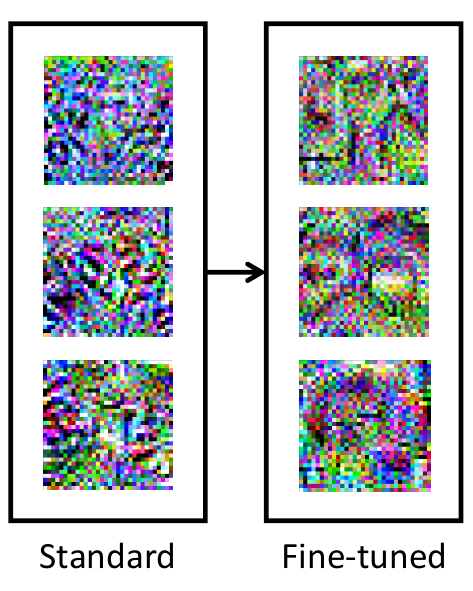}
    \vspace{0.05em}
    \end{minipage}
  }
  \caption{Generated low-entropy samples of standard and fine-tuned discriminators. For CIFAR10 and SVHN, the predicted labels of the selected low-entropy samples are ``Dog'' and ``Digit 6'', respectively.}
  \label{fig:GenOOD}
\end{figure}

\subsubsection{Generated Low-entropy Samples}
Fig.~\ref{fig:GenOOD} visualizes the generated low-entropy samples corresponding to standard discriminators and their fine-tuned discriminators. To generate low-entropy samples with more details from implicit generators, we set an extensive maximum iteration number $T = 10,000$ in LDS without an early stop. We adopt network architecture ResNet18. The presented low-entropy samples are labelled ``Dog'' and ``Digit 6'' on CIFAR10 and SVHN, respectively. We observe that the generated low-entropy samples by standard and fine-tuned discriminators are significantly different. For example, the samples generated by the fine-tuned discriminator contain more dark regions than that of its corresponding standard discriminator on CIFAR10. Furthermore, on SVHN, the generated samples of the standard discriminator are irregular, while the samples generated by its corresponding fine-tuned discriminator contain dark lines to form blurry digit 6. Recall that GAN generates visually meaningful samples by constraining the generated samples and training samples to satisfy the same distribution. Without this constraint, as shown in Fig.~\ref{fig:GenOOD}, the samples generated by implicit generators are unnecessarily consistent with real-world objects. Based on the discussion in Section~\ref{sec:ev}, the generated samples by standard and fine-tuned discriminators can be treated as the OOD samples with semantic shift and the ID samples with covariate shift, respectively.

\section{Conclusion and Future Work}
\label{sec:conclusion}
In this paper, we propose a method of \textit{fine-tuning discriminators by implicit generators} (FIG) to improve the OOD sensitivity of a given standard discriminator. FIG tackles the main challenge of generating discriminator-specific OOD samples. Specifically, we reveal the distributional vulnerability by the corresponding implicit generator inferred from a standard discriminator without extra training costs. A Langevin dynamic sampler draws OOD samples for the generator, which patches the distributional vulnerability by penalizing the prediction confidence of these generated samples. We empirically demonstrate that FIG outperforms the existing methods in detecting OOD samples.

The generated OOD samples address the distributional vulnerability. The training strategy of having OOD samples to patch the vulnerability and the results presented in this paper motivate a more ambitious direction: improving the classification generalization with OOD samples in training, i.e., classifying non-IID samples. This motivates the future task of making the involved OOD samples adaptive to both networks and training ID samples.


%


\ifCLASSOPTIONcompsoc
  \section*{Acknowledgments}
\else
  \section*{Acknowledgment}
\fi

The work is partially sponsored by Australian Research Council Discovery and Future Fellowship grants (DP190101079 and FT190100734).


\ifCLASSOPTIONcaptionsoff
  \newpage
\fi



%


\bibliographystyle{IEEEtran}
\bibliography{reffig}

\begin{thebibliography}{10}
\providecommand{\url}[1]{#1}
\csname url@samestyle\endcsname
\providecommand{\newblock}{\relax}
\providecommand{\bibinfo}[2]{#2}
\providecommand{\BIBentrySTDinterwordspacing}{\spaceskip=0pt\relax}
\providecommand{\BIBentryALTinterwordstretchfactor}{4}
\providecommand{\BIBentryALTinterwordspacing}{\spaceskip=\fontdimen2\font plus
\BIBentryALTinterwordstretchfactor\fontdimen3\font minus
  \fontdimen4\font\relax}
\providecommand{\BIBforeignlanguage}[2]{{%
\expandafter\ifx\csname l@#1\endcsname\relax
\typeout{** WARNING: IEEEtran.bst: No hyphenation pattern has been}%
\typeout{** loaded for the language `#1'. Using the pattern for}%
\typeout{** the default language instead.}%
\else
\language=\csname l@#1\endcsname
\fi
#2}}
\providecommand{\BIBdecl}{\relax}
\BIBdecl

\bibitem{RES:16}
K.~He, X.~Zhang, S.~Ren, and J.~Sun, ``Deep residual learning for image
  recognition,'' in \emph{CVPR}, 2016, pp. 770--778.

\bibitem{NN:19}
Z.~Allen{-}Zhu, Y.~Li, and Y.~Liang, ``Learning and generalization in
  overparameterized neural networks, going beyond two layers,'' in
  \emph{NeurIPS}, 2019, pp. 6155--6166.

\bibitem{CaoY022}
L.~Cao, P.~S. Yu, and Z.~Zhao, ``Shallow and deep non-iid learning on complex
  data,'' in \emph{{KDD}}.\hskip 1em plus 0.5em minus 0.4em\relax {ACM}, 2022,
  pp. 4774--4775.

\bibitem{Cao14}
L.~Cao, ``Non-iidness learning in behavioral and social data,'' \emph{Comput.
  J.}, vol.~57, no.~9, pp. 1358--1370, 2014.

\bibitem{HC:15}
A.~M. Nguyen, J.~Yosinski, and J.~Clune, ``Deep neural networks are easily
  fooled: High confidence predictions for unrecognizable images,'' in
  \emph{CVPR}, 2015, pp. 427--436.

\bibitem{OOD:21}
J.~Yang, K.~Zhou, Y.~Li, and Z.~Liu, ``Generalized out-of-distribution
  detection: {A} survey,'' \emph{CoRR}, pp. 1--20, 2021.

\bibitem{OOD:19}
J.~Ren, P.~J. Liu, E.~Fertig, J.~Snoek, R.~Poplin, M.~A. DePristo, J.~V.
  Dillon, and B.~Lakshminarayanan, ``Likelihood ratios for out-of-distribution
  detection,'' in \emph{NeurIPS}, 2019, pp. 14\,680--14\,691.

\bibitem{BL:17}
D.~Hendrycks and K.~Gimpel, ``A baseline for detecting misclassified and
  out-of-distribution examples in neural networks,'' in \emph{ICLR}, 2017, pp.
  1--12.

\bibitem{AT:17}
A.~Shrivastava, T.~Pfister, O.~Tuzel, J.~Susskind, W.~Wang, and R.~Webb,
  ``Learning from simulated and unsupervised images through adversarial
  training,'' in \emph{Conference on Computer Vision and Pattern Recognition},
  2017, pp. 2242--2251.

\bibitem{SF:16}
D.~Amodei, C.~Olah, J.~Steinhardt, P.~F. Christiano, J.~Schulman, and
  D.~Man{\'{e}}, ``Concrete problems in {AI} safety,'' \emph{CoRR}, pp. 1--29,
  2016.

\bibitem{UN:17}
A.~Kendall and Y.~Gal, ``What uncertainties do we need in {Bayesian} deep
  learning for computer vision?'' in \emph{NeurIPS}, 2017, pp. 5574--5584.

\bibitem{PN:18}
A.~Malinin and M.~J.~F. Gales, ``Predictive uncertainty estimation via prior
  networks,'' in \emph{NeurIPS}, 2018, pp. 7047--7058.

\bibitem{FIX:20}
H.~Touvron, A.~Vedaldi, M.~Douze, and H.~J{\'{e}}gou, ``Fixing the train-test
  resolution discrepancy: Fixefficientnet,'' in \emph{CoRR}, vol.
  abs/2003.08237, 2020, pp. 1--5.

\bibitem{RE:13}
Y.~Bengio, A.~C. Courville, and P.~Vincent, ``Representation learning: {A}
  review and new perspectives,'' \emph{TPAMI}, vol.~35, no.~8, 2013.

\bibitem{MIXUP:18}
H.~Zhang, M.~Ciss{\'{e}}, Y.~N. Dauphin, and D.~Lopez{-}Paz, ``mixup: Beyond
  empirical risk minimization,'' in \emph{ICLR}, 2018, pp. 1--13.

\bibitem{DA:19}
E.~D. Cubuk, B.~Zoph, D.~Mane, V.~Vasudevan, and Q.~V. Le, ``Autoaugment:
  Learning augmentation strategies from data,'' in \emph{CVPR}, 2019, pp.
  113--123.

\bibitem{PIC:17}
T.~Salimans, A.~Karpathy, X.~Chen, and D.~P. Kingma, ``Pixelcnn++: Improving
  the pixelcnn with discretized logistic mixture likelihood and other
  modifications,'' in \emph{ICLR}, 2017, pp. 1--10.

\bibitem{GLOW:18}
D.~P. Kingma and P.~Dhariwal, ``Glow: Generative flow with invertible 1x1
  convolutions,'' in \emph{NeurIPS}, 2018, pp. 10\,236--10\,245.

\bibitem{GSN:14}
Y.~Bengio, E.~Laufer, G.~Alain, and J.~Yosinski, ``Deep generative stochastic
  networks trainable by backprop,'' in \emph{ICML}, 2014, pp. 226--234.

\bibitem{MINE:18}
M.~I. Belghazi, A.~Baratin, S.~Rajeswar, S.~Ozair, Y.~Bengio, R.~D. Hjelm, and
  A.~C. Courville, ``Mutual information neural estimation,'' in \emph{ICML},
  2018, pp. 530--539.

\bibitem{EBM:06}
Y.~LeCun, S.~Chopra, R.~Hadsell, M.~Ranzato, and F.~J. Huang, ``A tutorial on
  energy-based learning,'' \emph{Predicting structured data}, vol.~1, no.~0,
  pp. 1--59, 2006.

\bibitem{SGLD:11}
M.~Welling and Y.~W. Teh, ``Bayesian learning via stochastic gradient langevin
  dynamics,'' in \emph{ICML}, 2011, pp. 681--688.

\bibitem{Cao-rs-eng}
L.~Cao, ``Non-iid recommender systems: A review and framework of recommendation
  paradigm shifting,'' \emph{Engineering}, vol.~2, no.~2, pp. 212--224, 2016.

\bibitem{Pang21}
G.~Pang, C.~Shen, L.~Cao, and A.~V.~D. Hengel, ``Deep learning for anomaly
  detection: A review,'' \emph{ACM Computing Surveys}, vol.~54, no.~2, pp.
  1--38, 2021.

\bibitem{Scheirer12}
W.~J. Scheirer, A.~de~Rezende~Rocha, A.~Sapkota, and T.~E. Boult, ``Toward open
  set recognition,'' \emph{IEEE Transactions on Pattern Analysis and Machine
  Intelligence}, vol.~35, no.~7, pp. 1757--1772, 2013.

\bibitem{MacedoRZOL22}
D.~Mac{\^{e}}do, T.~I. Ren, C.~Zanchettin, A.~L.~I. Oliveira, and T.~B.
  Ludermir, ``Entropic out-of-distribution detection: Seamless detection of
  unknown examples,'' \emph{{IEEE} Trans. Neural Networks Learn. Syst.},
  vol.~33, no.~6, pp. 2350--2364, 2022.

\bibitem{BahriSPS22}
M.~Bahri, F.~Salutari, A.~Putina, and M.~Sozio, ``Automl: state of the art with
  a focus on anomaly detection, challenges, and research directions,''
  \emph{Int. J. Data Sci. Anal.}, vol.~14, no.~2, pp. 113--126, 2022.

\bibitem{XiangWRSDZ21}
H.~Xiang, J.~Wang, K.~Ramamohanarao, Z.~Salcic, W.~Dou, and X.~Zhang,
  ``Isolation forest based anomaly detection framework on non-iid data,''
  \emph{{IEEE} Intell. Syst.}, vol.~36, no.~3, pp. 31--40, 2021.

\bibitem{Chen22}
G.~Chen, P.~Peng, X.~Wang, and Y.~Tian, ``Adversarial reciprocal points
  learning for open set recognition,'' \emph{IEEE Transactions on Pattern
  Analysis and Machine Intelligence}, vol. DOI 10.1109/TPAMI.2021.3106743,
  2021.

\bibitem{Li-etal21}
H.~Li, X.~Wang, Z.~Zhang, and W.~Zhu, ``{OOD-GNN:} out-of-distribution
  generalized graph neural network,'' \emph{IEEE Transactions on Knowledge and
  Data Engineering}, vol. DOI 10.1109/TKDE.2022.3193725, 2021.

\bibitem{KD:15}
G.~E. Hinton, O.~Vinyals, and J.~Dean, ``Distilling the knowledge in a neural
  network,'' in \emph{CoRR}, vol. abs/1503.02531, 2015.

\bibitem{ODIN:18}
S.~Liang, Y.~Li, and R.~Srikant, ``Enhancing the reliability of
  out-of-distribution image detection in neural networks,'' in \emph{ICLR},
  2018, pp. 1--27.

\bibitem{FC:21}
J.~van Amersfoort, L.~Smith, A.~Jesson, O.~Key, and Y.~Gal, ``Improving
  deterministic uncertainty estimation in deep learning for classification and
  regression,'' in \emph{CoRR}, 2021, pp. 1--16.

\bibitem{MLB:18}
K.~Lee, K.~Lee, H.~Lee, and J.~Shin, ``A simple unified framework for detecting
  out-of-distribution samples and adversarial attacks,'' in \emph{NeurIPS},
  2018, pp. 7167--7177.

\bibitem{DFR:20}
E.~Zisselman and A.~Tamar, ``Deep residual flow for out of distribution
  detection,'' in \emph{CVPR}, 2020, pp. 13\,991--14\,000.

\bibitem{GM:20}
C.~S. Sastry and S.~Oore, ``Detecting out-of-distribution examples with gram
  matrices,'' in \emph{ICML}, 2020, pp. 8491--8501.

\bibitem{DGM:19}
E.~T. Nalisnick, A.~Matsukawa, Y.~W. Teh, D.~G{\"{o}}r{\"{u}}r, and
  B.~Lakshminarayanan, ``Do deep generative models know what they don't know?''
  in \emph{ICLR}, 2019, pp. 1--19.

\bibitem{LR:19}
J.~Ren, P.~J. Liu, E.~Fertig, J.~Snoek, R.~Poplin, M.~A. DePristo, J.~V.
  Dillon, and B.~Lakshminarayanan, ``Likelihood ratios for out-of-distribution
  detection,'' in \emph{NeurIPS}, 2019, pp. 14\,680--14\,691.

\bibitem{LGM:20}
J.~Serr{\`{a}}, D.~{\'{A}}lvarez, V.~G{\'{o}}mez, O.~Slizovskaia, J.~F.
  N{\'{u}}{\~{n}}ez, and J.~Luque, ``Input complexity and out-of-distribution
  detection with likelihood-based generative models,'' in \emph{ICLR}, 2020,
  pp. 1--15.

\bibitem{OE:19}
D.~Hendrycks, M.~Mazeika, and T.~G. Dietterich, ``Deep anomaly detection with
  outlier exposure,'' in \emph{ICLR}, 2019, pp. 1--18.

\bibitem{SSS:19}
P.~Bevandic, I.~Kreso, M.~Orsic, and S.~Segvic, ``Simultaneous semantic
  segmentation and outlier detection in presence of domain shift,'' in
  \emph{PRGC}, 2019, pp. 33--47.

\bibitem{FH:19}
H.~Blum, P.~Sarlin, J.~I. Nieto, R.~Siegwart, and C.~Cadena, ``Fishyscapes: {A}
  benchmark for safe semantic segmentation in autonomous driving,'' in
  \emph{ICCVW}, 2019, pp. 2403--2412.

\bibitem{DCC:20}
Y.~Hsu, Y.~Shen, H.~Jin, and Z.~Kira, ``Generalized {ODIN} detecting
  out-of-distribution image without learning from out-of-distribution data,''
  in \emph{CVPR}, 2020, pp. 10\,948--10\,957.

\bibitem{AD:15}
I.~J. Goodfellow, J.~Shlens, and C.~Szegedy, ``Explaining and harnessing
  adversarial examples,'' in \emph{ICLR}, 2015, pp. 1--11.

\bibitem{GO:18}
K.~Lee, H.~Lee, K.~Lee, and J.~Shin, ``Training confidence-calibrated
  classifiers for detecting out-of-distribution samples,'' in \emph{ICLR},
  2018, pp. 1--16.

\bibitem{IG:18}
W.~Grathwohl, K.~Wang, J.~Jacobsen, D.~Duvenaud, M.~Norouzi, and K.~Swersky,
  ``Your classifier is secretly an energy based model and you should treat it
  like one,'' in \emph{ICLR}, 2020, pp. 1--23.

\bibitem{VB:19}
B.~Poole, S.~Ozair, A.~van~den Oord, A.~Alemi, and G.~Tucker, ``On variational
  bounds of mutual information,'' in \emph{ICML}, 2019, pp. 5171--5180.

\bibitem{MCMC:17}
R.~Bardenet, A.~Doucet, and C.~C. Holmes, ``On {Markov Chain Monte Carlo}
  methods for tall data,'' \emph{J. Mach. Learn. Res.}, vol.~18, pp. 1--47,
  2017.

\bibitem{GI:02}
G.~E. Hinton, ``Training products of experts by minimizing contrastive
  divergence,'' \emph{Neural Comput.}, vol.~14, no.~8, pp. 1771--1800, 2002.

\bibitem{IG:19}
Y.~Du and I.~Mordatch, ``Implicit generation and modeling with energy based
  models,'' in \emph{NeurIPS}, 2019, pp. 3603--3613.

\bibitem{ML:14}
S.~Shalev-Shwartz and S.~Ben-David, \emph{Understanding Machine Learning From
  Theory to Algorithms}.\hskip 1em plus 0.5em minus 0.4em\relax Cambridge
  University Press, 2014.

\bibitem{PGD:18}
A.~Madry, A.~Makelov, L.~Schmidt, D.~Tsipras, and A.~Vladu, ``Towards deep
  learning models resistant to adversarial attacks,'' in \emph{ICLR}, 2018, pp.
  1--23.

\bibitem{GAN:14}
I.~J. Goodfellow, J.~Pouget{-}Abadie, M.~Mirza, B.~Xu, D.~Warde{-}Farley,
  S.~Ozair, A.~C. Courville, and Y.~Bengio, ``Generative adversarial nets,'' in
  \emph{NeurIPS}, 2014, pp. 2672--2680.

\bibitem{CP:17}
G.~Pereyra, G.~Tucker, J.~Chorowski, L.~Kaiser, and G.~E. Hinton,
  ``Regularizing neural networks by penalizing confident output
  distributions,'' in \emph{ICLR}, 2017, pp. 1--11.

\bibitem{MC:19}
T.~Wu and D.~F. Gleich, ``Multiway {Monte Carlo} method for linear systems,''
  \emph{{SIAM} J. Sci. Comput.}, vol.~41, no.~6, pp. 3449--3475, 2019.

\bibitem{SVHN:11}
Y.~Netzer, T.~Wang, A.~Coates, A.~Bissacco, B.~Wu, and A.~Y. Ng, ``Reading
  digits in natural images with unsupervised feature learning,'' in \emph{NIPS
  Workshop on Deep Learning and Unsupervised Feature Learning}, 2011.

\bibitem{CIFAR10:09}
A.~Krizhevsky, ``Learning multiple layers of features from tiny images,'' Tech.
  Rep., 2009.

\bibitem{OSL:16}
O.~Vinyals, C.~Blundell, T.~Lillicrap, K.~Kavukcuoglu, and D.~Wierstra,
  ``Matching networks for one shot learning,'' in \emph{NeurIPS}, 2016, pp.
  3630--3638.

\bibitem{LSUN:15}
F.~Yu, Y.~Zhang, S.~Song, A.~Seff, and J.~Xiao, ``{LSUN:} construction of a
  large-scale image dataset using deep learning with humans in the loop,''
  \emph{CoRR}, vol. abs/1506.03365, 2015.

\bibitem{IMAGENET:09}
J.~Deng, W.~Dong, R.~Socher, L.~Li, K.~Li, and F.~Li, ``Imagenet: {A}
  large$-$scale hierarchical image database,'' in \emph{CVPR}, 2009, pp.
  248--255.

\bibitem{CAL:06}
G.~Griffin, A.~Holub, and P.~Perona, ``The {Caltech} 256,'' Caltech Technical
  Report, Tech. Rep., 2006.

\bibitem{COCO:14}
T.~Lin, M.~Maire, S.~J. Belongie, J.~Hays, P.~Perona, D.~Ramanan,
  P.~Doll{\'{a}}r, and C.~L. Zitnick, ``Microsoft {COCO}: Common objects in
  context,'' in \emph{ECCV}, vol. 8693, 2014, pp. 740--755.

\bibitem{Oxfordflowers102}
M.~Nilsback and A.~Zisserman, ``A visual vocabulary for flower
  classification,'' in \emph{{IEEE} Computer Society Conference on Computer
  Vision and Pattern}, no. 1447--1454, 2006.

\bibitem{DTD47}
M.~Cimpoi, S.~Maji, I.~Kokkinos, S.~Mohamed, , and A.~Vedaldi, ``Describing
  textures in the wild,'' in \emph{{IEEE} Conference on Computer Vision and
  Pattern Recognition}, no. 3606--3613, 2014.

\bibitem{VGG:15}
C.~Szegedy, W.~Liu, Y.~Jia, P.~Sermanet, S.~E. Reed, D.~Anguelov, D.~Erhan,
  V.~Vanhoucke, and A.~Rabinovich, ``Going deeper with convolutions,'' in
  \emph{CVPR}, 2015, pp. 1--9.

\bibitem{SHU:18}
N.~Ma, X.~Zhang, H.~Zheng, and J.~Sun, ``Shufflenet {V2:} practical guidelines
  for efficient {CNN} architecture design,'' in \emph{ECCV}, 2018, pp.
  122--138.

\bibitem{DEN:17}
G.~Huang, Z.~Liu, L.~van~der Maaten, and K.~Q. Weinberger, ``Densely connected
  convolutional networks,'' in \emph{CVPR}, 2017, pp. 2261--2269.

\bibitem{MS:18}
G.~Shalev, Y.~Adi, and J.~Keshet, ``Out-of-distribution detection using
  multiple semantic label representations,'' in \emph{NeurIPS}, 2018, pp.
  7386--7396.

\bibitem{TP:21}
H.~Touvron, M.~Cord, M.~Douze, F.~Massa, A.~Sablayrolles, and H.~J{\'{e}}gou,
  ``Training data-efficient image transformers {\&} distillation through
  attention,'' in \emph{ICML}, 2021, pp. 10\,347--10\,357.

\bibitem{SNE:08}
L.~van~der Maaten and G.~Hinton, ``Visualizing data using t-sne,'' \emph{JMLR},
  vol.~9, no.~86, pp. 2579--2605, 2008.

\end{thebibliography}

%

\begin{IEEEbiography}[{\includegraphics[width=1in,height=1.25in,clip,keepaspectratio]{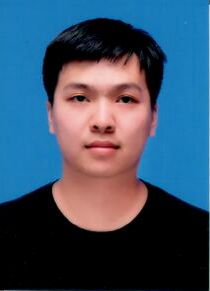}}]{Zhilin Zhao}
received a B.S. and M.S. degree from the School of Data and Computer Science, Sun Yat-Sen University, China. He is currently a PhD student in the School of Computer Science, University of Technology Sydney, Australia. His research interests include generalization analysis, online learning, and out-of-distribution detection.
\end{IEEEbiography}


\begin{IEEEbiography}[{\includegraphics[width=1in,height=1.25in,clip,keepaspectratio]{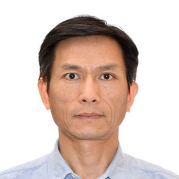}}]{Longbing Cao} 
is a Professor at the University of Technology Sydney and an ARC Future Fellow (Level 3). He received one PhD in Pattern Recognition and Intelligent Systems from the Chinese Academy of Sciences and another in Computing Science at UTS. His research interests include artificial intelligence, data science, knowledge discovery, machine learning, behavior informatics, complex intelligent systems, and enterprise innovation.
\end{IEEEbiography}


\begin{IEEEbiography}[{\includegraphics[width=1in,height=1.25in,clip,keepaspectratio]{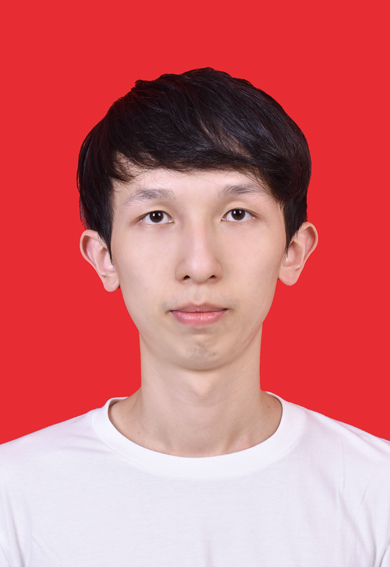}}]{Kun-Yu Lin}
received a B.S. and M.S. degree from the School of Data and Computer Science, Sun Yat-sen University, China. He is currently a PhD student in the School of Computer Science and Engineering, Sun Yat-sen University. His research interests include computer vision and machine learning.
\end{IEEEbiography}





\end{document}